\documentclass[11pt]{article}

\usepackage[preprint]{acl}

\usepackage{times}
\usepackage{latexsym}

\usepackage[T1]{fontenc}

\usepackage[utf8]{inputenc}

\usepackage{microtype}

\usepackage{inconsolata}

\usepackage{graphicx}
\usepackage[dvipsnames]{xcolor}

\usepackage{booktabs}
\usepackage{tabularx}
\usepackage{enumerate}
\usepackage{enumitem}
\setlist[enumerate]{
  topsep=4pt,
  itemsep=2pt,
  parsep=0pt,
  partopsep=0pt
}
\usepackage{helvet}
\usepackage{amsmath}
\usepackage{amssymb}
\usepackage{subcaption}

\usepackage{mdframed}
\newmdenv[
  topline=false,
  bottomline=false,
  rightline=false,
  linecolor=gray,
  linewidth=1pt,
  skipabove=\topsep,
  skipbelow=\topsep
]{draftytext}

\title{On Emergent Social World Models --- Evidence for Functional Integration of Theory of Mind and Pragmatic Reasoning in Language Models}

\author{Polina Tsvilodub$^*$ \and Jan-Felix Klumpp \and Amir Mohammadpour \\
        Department of Linguistics \\
        University of T\"ubingen \\
        \texttt{
           \texttt{$^*$polina.tsvilodub@uni-tuebingen.de}
         }
        \AND
        Jennifer Hu\\Department of Cognitive Science \\ Johns Hopkins University
        \And 
        Michael Franke \\ Department of Linguistics \\
        University of T\"ubingen
        }

\begin{document}
\maketitle
\begin{abstract}
This paper investigates whether LMs recruit shared computational mechanisms for general Theory of Mind (ToM) and language-specific pragmatic reasoning in order to contribute to the general question of whether LMs may be said to have emergent ``social world models,'' i.e., representations of mental states that are repurposed across tasks (the \textit{functional integration hypothesis}). 
Using behavioral evaluations and causal-mechanistic experiments via functional localization methods inspired by cognitive neuroscience, we analyze LMs' performance across seven subcategories of ToM abilities \citep{beaudoin2020systematic} on a substantially larger localizer dataset than used in prior like-minded work.
Results from stringent hypothesis-driven statistical testing offer suggestive evidence for the functional integration hypothesis, indicating that LMs may develop interconnected ``social world models'' rather than isolated competencies. 
This work contributes novel ToM localizer data, methodological refinements to functional localization techniques, and empirical insights into the emergence of social cognition in artificial systems.
\end{abstract}

\section{Introduction}
\label{sec:introduction}

Large language models (LMs) possess astonishing abilities and prove useful for a plethora of downstream tasks, but controversy persists regarding how to conceptualize their capacities.
Some view LMs as mere ``stochastic parrots,'' fundamentally incapable of genuine understanding \cite{BenderKoller2020:Climbing-toward,BenderGebru2021:On-the-Dangers-}. 
Others argue that highly proficient language ability presupposes the emergence of ``world models,'' i.e., representations of (relevant aspects of) the language-generating processes \cite{Yudkowsky2023:GPTs-are-Predic,BereskaGavves2024:Mechanistic-Int}. 
Arguments for the latter view typically come from models optimized for solving finite-state problems like Chess \cite{ToshniwalWiseman2022:Chess-as-a-Test}, Othello \cite{LiHopkins2023:Emergent-World-}, or taxi navigation \cite{VafaChen2024:Evaluating-the-}, or from narrow domains like text-based adventure games \cite{LiNye2021:Implicit-repres}.
But to what extent might large pretrained LMs recruit representations and computational mechanisms suggestive of ``world models'' for tasks crucial for \textit{general-purpose language understanding}?

Language understanding comprises \emph{formal} and \emph{functional} abilities \cite{MahowaldIvanova2024:Dissociating-la}. 
Formal abilities (e.g., syntax) may be acquired relatively straightforwardly from local statistical patterns in training data. 
Functional abilities relate to language use in communication, e.g., \textit{pragmatic reasoning}, such as comprehension of contextual meaning and social implications of utterances \cite{LevinsonPragmatics1983}.
Functional ability seems more difficult to extract from text alone, since the surface language signal provides less direct information about relevant latent variables, such as interlocutors' mental states.
This raises the issue: to what extent do LMs develop rich ``social world models''---entertaining concepts of self and other, social commitment, and nested beliefs---in order to achieve their observed level of performance?

We here address this big-picture question with a concrete case study on the relationship between LMs' capacities for (i) Theory of Mind reasoning and (ii) pragmatic reasoning. 
Theory of Mind (ToM) refers to the capability to attribute mental states---unobservable, latent properties such as beliefs or desires---to oneself and other agents, and plays a fundamental role in social cognition \citep{beaudoin2020systematic} and pragmatic language use \citep{Grice1975:Logic-and-Conve, goodman2016pragmatic}.
Evidence from cognitive neuroscience indicates that humans indeed have, at least to some extent, shared functional mechanisms, namely, overlapping neural substrates supporting both Theory of Mind and pragmatic reasoning \cite{SaxeKanwisher2003:People-thinking,bosco2017neural,hauptman2023non}.

What holds for human cognition need not, of course, generalize to LMs. 
Nothing in their architecture explicitly enforces structural similarity to human solutions for functional language abilities. 
Nevertheless, if shared computational mechanisms govern linguistic and non-linguistic social reasoning in humans, efficient compression principles might favor cross-purposing internal mechanisms across these domains in stochastic learners as well. 
We therefore formulate two contrasting hypotheses.
The \textbf{functional specialization} hypothesis holds that LMs use different computational mechanisms for pragmatic and ToM-reasoning.
In contrast, the \textbf{functional integration} hypothesis maintains that computational mechanisms for pragmatic and ToM-reasoning are shared, at least in substantial part.

We explore whether LMs exhibit functional specialization or functional integration through evidence from behavioral evaluations and causal-mechanistic experiments.
The latter builds on recent work by \citet{alkhamissi-etal-2025-llm} on \textit{functional localization} inspired from cognitive neuroscience which identifies subsets of neurons in LMs supporting performance on target tasks \cite[cf.,][]{duan-etal-2025-unveiling,hanna2025formal}.
While their work did not find strong evidence for functional ToM localization based on a very small data set, we do find causal evidence for it based on a much larger data set which is part of this paper's contribution.

The two hypotheses structure our investigation, even if this dichotomy is a simplification in several respects. 
The repurposing of computational mechanisms likely exists along a continuum rather than as a binary distinction, and the constructed domains of pragmatic and ToM-reasoning may themselves lack internal homogeneity. 
To acknowledge this general question of the level of granularity at which function sharing should be investigated, we investigate LMs' ToM capabilities while taking into account the ``Aspects in Theory Of Mind Space'' (ATOMS) framework.
ATOMS identifies seven fine-grained subcategories in ToM abilities (\citealp{beaudoin2020systematic}; beliefs, intentions, desires, emotions, knowledge, percepts, mentalistic understanding of non-literal communication; described in detail in Appendix~\ref{sec:app:atoms}). 
We conduct evaluations of LMs on datasets tapping into different ATOMS, building on like-minded work \cite{ma-etal-2023-towards-holistic}.


\paragraph{Related Work \& Contributions}
Our paper builds on ideas from numerous studies in psychology and cognitive science, 
showing that Theory of Mind enables humans to interpret others' actions as causally driven by these mental states \citep[e.g.,][]{beaudoin2020systematic, goodman2006intuitive, wimmer1983beliefs}.
A longstanding tradition in philosophy, linguistics, and cognitive science conceptualizes communication as a specialized form of goal-directed action \cite{austin1975things, Grice1975:Logic-and-Conve},
such that understanding language 
is, in principle, structurally similar to interpreting non-linguistic actions, and has also been conceptualized as interlocutors engaging in probabilistic recursive reasoning about each other's mental states \citep{Grice1975:Logic-and-Conve, goodman2016pragmatic}. 

Additionally, we contribute to related work evaluating LMs, often separately focusing on ToM and linguistic abilities, in the domain of mechanistic interpretability \citep{wang2022interpretability}, in particular LMs' language capabilities \citep{oh2025tug} and mental state representations \citep{bortoletto2024brittle, zhu2024language}, and to the growing number of behavioral evaluations of LMs' ToM and pragmatic abilities \citep[which, in part, provided mixed results; e.g.,][]{hu-etal-2023-fine, ullman2023large, sap-etal-2022-neural, ma-etal-2025-pragmatics}.
While the functional connection of these two abilities has been addressed more in human research \citep{baldwin_language_2005, hauptman2023non}, work on the functional organization of LMs has mostly focused on aligning only LMs' language representations to human brains \citep{tuckute2024language, alkhamissi-etal-2025-language}.
We follow recent arguments for more integrated investigations of linguistic abilities of LMs grounded in cognitive science approaches \citep{sumers2023cognitive, mccoy2024embers}.

In sum, this paper contributes behavioral and causal-mechanistic experiments (Sections~\ref{sec:behavioral-eval} \& \ref{sec:functional-localization} respectively), paired with stringent hypothesis-driven statistical testing, that provide suggestive evidence in favor of the functional integration view, thus contributing to the question of emergent social world models.
In doing so, the paper contributes novel, empirically grounded ToM localizer data and methodological refinements of the functional localization method.\footnote{We release all code and data at \url{https://github.com/polina-tsvilodub/lm-emergent-social-world-models}.}

\section{Behavioral Evaluation}
\label{sec:behavioral-eval}

According to the functional integration hypothesis, if ToM and pragmatic reasoning share computational resources, we expect a systematic relation between performance on these 
tasks across different LMs, which we operationalize in three predictions to be tested by statistical analyses:

\begin{enumerate}
    \item[\textbf{P1}] \textbf{(simple correlation):}
    Across LMs, accuracies on ToM and pragmatic reasoning are positively correlated.
    
    \item[\textbf{P2}] \textbf{(predictive redundancy)}:
    Under appropriate controls, when predicting accuracy for a task, information about whether a task is a pragmatic or a ToM reasoning task is not relevant for predicting an LMs' accuracy for it.
    
    \item[\textbf{P3}] \textbf{(predictive gains)}:
    Under appropriate controls, information on an LM's ToM accuracy is better for predicting its pragmatic accuracy than information on general language ability.
    
\end{enumerate}

While P1--P2 address the behavioral correlation between ToM and pragmatic capabilities, P3 helps to disentangle this connection from potential confounds: both ToM and pragmatic capabilities likely correlate with general language ability (e.g., for instruction comprehension), as well as on general cognitive abilities and world knowledge. 
While the following analyses focus on controlling for general linguistic ability through performance on two linguistic benchmarks, we provide additional exploratory mechanistic analyses of the models also on general intelligence tasks like analogical reasoning and entity tracking in Section~\ref{sec:ablation-experiments} and in Appendix~\ref{app:sec:ablations-g-laoding-datasets}.

\subsection{Datasets}
\label{sec:eval-datasets}
We select 16 datasets testing ten different pragmatic phenomena from extant pragmatics benchmarks \citep{ma-etal-2025-pragmatics}, and 22 datasets from existing ToM benchmarks. 
We use several ToM benchmarks with ATOMS annotations introduced by \citet{ma-etal-2023-towards-holistic}, 
and select additional ToM datasets, annotating them with ATOMS we consider the benchmarks to likely test. 
All datasets have a multiple-choice format: for each item $x$, given a context $c$ that contains the target vignette, a question, and optionally instructions or the answer options, there are $2 \leq n \leq 6$ answer options, one of which is the  correct answer that is consistent with ToM-based or pragmatic reasoning.
For evaluating general language ability of each model, the BLiMP dataset \citep{warstadt-etal-2020-blimp-benchmark} and the SNLI dataset \citep{bowman2015large} are used.
The detailed list of datasets and annotations can be found in Appendix~\ref{sec:app:behavioral-eval-datasets}, full prompts for each dataset are reported in Appendix~\ref{sec:app:eval-dataset-formatting}.

\subsection{Models \& Procedure}
\label{sec:behavioral-eval-models}
We test 48 open-weights and/or open-source LMs from seven families (Llama, Qwen, Falcon, Mistral, OLMo, Pythia, Gemma), ranging from 0.5B to 72B parameters (listed in Appendix~\ref{sec:app:behavioral-eval-models}). 
Where possible, we test base and fine-tuned versions, the type of which (instruction-tuning, RLHF, both) depends on the model.

LMs' predictions are retrieved via conditional log-probability scoring. 
For each item, we retrieve the log-probability of the tokens of each answer option, conditioned on the input prompt, and average across the token log-probabilities to correct for different lengths of different answer options. 
The model's prediction is the option with the highest normalized log-probability.
Accuracy is the proportion of correct answers. 

\subsection{Results}
\label{sec:results:behavioral-evals}

\begin{figure}[t!]
\centering
    \includegraphics[width=\linewidth]{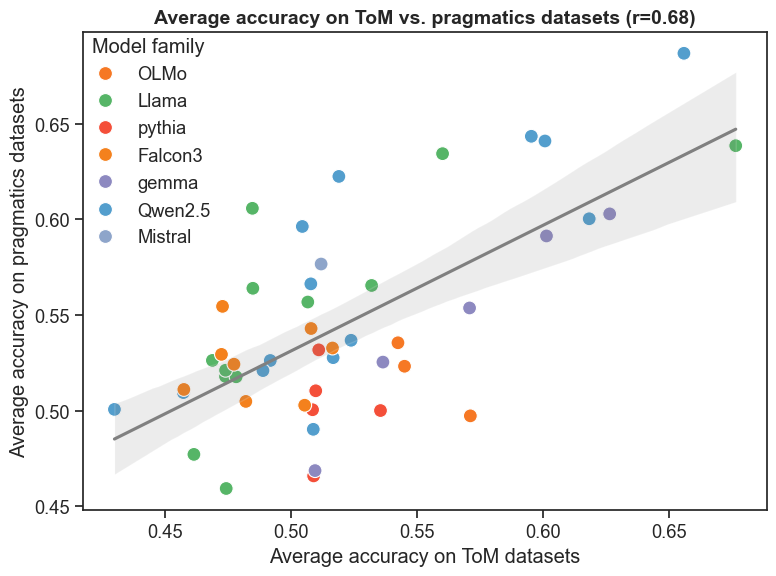}
    \caption{Average accuracy of 48 models on ToM (x-axis) and pragmatics (y-axis) tasks. \label{fig:acc-tom-vs-prag}}
\end{figure}

Figure~\ref{fig:acc-tom-vs-prag} shows the correlation between models' accuracy on pragmatic and ToM benchmarks.
As per prediction P1, we find a moderate but significant positive correlation ($r = 0.68$, $p=1.24 \cdot 10^{-7}$).

\bigskip


To address prediction P2, we fit a Bayesian beta regression model in R using the \texttt{brms} package \citep{burkner2017brms}, regressing the average accuracy of each model on each dataset against the main effects of model family (seven levels), model size (where models up to 8B parameters were labeled as small, up to 32B as medium, and up to 72B as large), model type (base~vs.~fine-tuned), dataset type (instruction / answer options in context~vs.~sentence completion) and domain (ToM~vs.~pragmatics), an interaction between model size and type, and random by-dataset intercepts.\footnote{The regression model in R syntax: \texttt{accuracy $\sim$ model\_family + size * model\_type + ds\_type + domain + (1 | dataset)}. A model with full interaction structure did not converge. 
} Here and below, we report posterior means and 95\% credible intervals for the posterior of contrasts of interest. All independent variables were sum-coded.
In line with most evaluations in the literature, we found that, across domains, fine-tuned models performed credibly better than base models ($\beta=0.11 [0.05, 0.16]$), large models performed better than medium models ($\beta=0.20 [0.11, 0.29]$) and medium models performed better than small models ($\beta=0.26 [0.20, 0.32]$).
We found no differences between the dataset types ($\beta=-0.06 [-0.88, 0.73]$).
Crucially, in line with prediction P2, we also found no difference between the domains ToM and pragmatics ($\beta = -0.03 [-0.74, 0.67]$), suggesting that the information about the domain of a given task 
is not crucial for predicting a given model's accuracy if sufficient information about the model is given. 
Further accuracy results on the single datasets are in the Appendix (Figure~\ref{app:fig:prag-acc-part1}--Figure~\ref{app:fig:tom-acc-part2}).

Finally, 
we address P3 with model comparison for two Bayesian beta regression models predicting the average accuracy on pragmatics datasets for each model against the baseline predictors of model size 
and model type.\footnote{Including the model family fixed effect led to convergence issues.}
One model (M0) contains the additional predictor of the model's average accuracy on the linguistic benchmarks.
The other model (M1) contains information of the model's average accuracy on ToM tasks. 
Comparing M0 and M1 via leave-one-out cross-validation (LOO-CV), we found that model M1 was significantly better than M0, measured through expected log predictive density (ELPD) ($ELPD=-16.1, \; p\text{-value}=0$), thus supporting prediction P3.
Additional details on our statistical analyses are in Appendix~\ref{sec:app:behavioral-eval-stats}.

To explore potential diversity in ToM reasoning further, we also ask whether there are more specific \textit{aspects} of the ToM capabilities that LMs perform particularly well on.
We operationalize this question through our annotations of the evaluation datasets with the ATOMS framework \cite{beaudoin2020systematic}.
We compare the baseline beta regression model (MB) where a model's average accuracy on ToM evaluation datasets is regressed against the model family, size, type and dataset type, to models which additionally include all possible combinations of the seven ATOMS main effects, 
resulting in 128 different regression models.
Comparison was also conducted via LOO-CV.
We found that the baseline MB was numerically worse than any model with ATOMS predictors.
The best model additionally contained the aspects intentions, desires, emotions, percepts and non-literal communication.
The difference between MB and the best model was significant ($ELPD=-58.80\; p\text{-value}=0$; full results are in the Appendix, Figure~\ref{app:fig:atoms-model-comparison}).
Model comparison further suggests that several models are on par with the best model, and several models next to MB are clearly worse than others.
The least performing models do not contain the predictor percepts, while containing up to three additional other ATOMS predictors. 
This suggests that the factor percepts added the most predictive power to the statistical models of the LMs' accuracy, specifically distinguishing datasets which tapped into this ToM aspect (false-belief tasks about the location of an object and tasks about agent properties) from other evaluation datasets.
Conjointly, these behavioral results provide first evidence for the functional integration hypothesis. Next, we investigate whether the hypothesis is supported at the level of LMs' internal mechanisms.

\section{Functional Localization}
\label{sec:functional-localization}

Functional subnetwork localization is a method that has been widely used in cognitive neuroscience to identify regions (or, subnetworks) in the human brain that are implied in the execution of a particular task or ability, including Theory of Mind \cite[e.g.,][]{SaxeBrett2006:Divide-and-conq,fedorenko2010new}, and indeed \citet{hauptman2023non} suggest that pragmatic language use in humans relies on the same ToM subnetwork.
The procedure uses so-called \textit{localizer suites} that contrast a \textit{target} condition which contains stimuli that require the capacity of interest for processing, against a closely matched \textit{control} condition which does not require it.
For example, the ToM network has been localized by contrasting target stories in which an actor holds a false belief  \citep[][]{baron1985does} against control stories about pictures containing false information \citep[][]{Zaitchik1990:When-representa}.
While reading these stories, the brain activity of human participants is recorded and localization proceeds via by-individual identification of regions that are selectively active under the target, but not under the control stimuli.
%
However, given that ToM may itself not be a single monolithic capability (see Section~\ref{sec:introduction}), cognitive neuroscience has also used other localizers than false belief tasks, e.g., economic games or stories involving moral intent, which have been found to activate the same functional brain region in humans.
For our localizations, we build on recent work extending this procedure to language models \citep{alkhamissi-etal-2025-llm}, described next.

\begin{table*}[t]
    \centering
    \begin{tabularx}{\linewidth}{l X}
        \toprule
         %
         \multicolumn{2}{l}{\textbf{LatentBeliefs} \textcolor{gray}{(Example of GPT-5 generated stimuli based on data from \citealp{SaxeKanwisher2003:People-thinking})}} \\
         FalseBelief (trgt)  & Tom dropped his keys in the bowl by the door before going for a run. While he was out, Mia put the keys into the desk drawer to tidy up. Tom came back a bit later. S / Q: Tom will first look for his keys in the ($\bullet \text{ bowl} \bullet \text{ drawer}$) \\
         Desire (trgt)       & Coach Ramirez booked the soccer field for Saturday practice. He told the team they’d be outside unless lightning forced them into the gym. A storm never came, and they kept the plan. S / Q: The coach wanted to practice ($\bullet \text{ indoors} \bullet \text{ outdoors}$) \\
         FalsePhoto (ctrl)   & A bus schedule was printed last month. Yesterday the transit agency changed Route 9 to depart an hour later. All the old paper schedules are still circulating. S / Q: On the printed schedule, Route 9 departs ($\bullet \text{ earlier} \bullet \text{ later}$) \\
         \midrule
         \multicolumn{2}{l}{\textbf{GameBeliefs} \textcolor{gray}{(Example of GPT-5 generated stimuli based on data from \citealp{chang2023mentalizing})}} \\
         GameBelief (trgt)   & Nora and Ben play an Ultimatum Game. Nora is the proposer and receives 20 Euros to divide between her and another player. She intends to offer half to Ben, but a glitch records her offer as 1 Euro. Ben sees 1 Euro on his screen, frowns, and rejects immediately. S / Q: Ben evidently believes that ($\bullet \text{ Nora is trying to cheat him} \bullet \text{ a glitch reduced the offer}$) \\
         GameOutcome (ctrl)  & Nora and Ben play an Ultimatum Game. Nora is the proposer and receives 20 Euros to divide between her and another player. She intends to offer half to Ben, but a glitch records her offer as 1 Euro. Ben sees 1 Euro on his screen, frowns, and rejects immediately. S / Q: Ben will now have ($\bullet \text{ 1 Euro} \bullet \text{ 0 Euro}$) \\
        \bottomrule
    \end{tabularx}
    \caption{Example synthetic stimuli for five conditions from two localizer suites. Dots indicate answer options. ``S'' / ``Q:'' stand for ``Statement'' / ``question:'' used in the prompts. ``trgt'' indicates a target stimulus, ``ctrl'' a control stimulus. Examples for all suites are in the Appendix, Table~\ref{tab:LocalizerSuitesAll}.
    }
    \label{tab:LocalizerSuites}
\end{table*}

\subsection{Localizer Suites}
\label{sec:synthetic-stimuli}

A \textit{localizer suite} $L$ comprises a set of sets of target stimuli $\{ S_T^{(j)} \}_{1 \le j \le n_j}$ and a set of sets of control stimuli $\{ S_C^{(k)}\}_{1 \le k \le n_k}$. 
Localizer suites with $n_j = n_k = 1$ are called \textit{simple}.
We synthetically generated four localizer suites 
based on extant experimental material which was successful in previous fMRI studies for singling out the ToM network in human brains. 
These prior studies were selected for variety and to cover all of the seven top-level categories of the ATOMS framework, while still sticking with empirically validated material:\footnote{
    For example, belief and percepts are relevant for the LatentBelief suite, intentions and knowledge are relevant for the MoralIntent suite, desires and emotions are relevant for the GameBelief suite, and the CommunicativeIntent suite addresses non-literal communication.
} 
The \textbf{LatentBeliefs} suite covers the commonly used false-belief items \cite{SaxeKanwisher2003:People-thinking}, 
\textbf{CommunicativeIntent} looks at mental state ascriptions in communication, and might be particularly relevant for comparison to pragmatic reasoning \cite{bosco2017neural},
\textbf{GameBeliefs} considers ToM reasoning in the context of economic games \citep[trust \& ultimatum;][]{chang2023mentalizing}, and 
\textbf{MoralIntent} considers issues of responsibility in decision making \cite{young2007neural}.
Our localizer suites contain different numbers of target and control sets (each called condition): 
the two simple suites (GameBeliefs and MoralIntent) have target and control sets in which items are exactly paired, two suites (LatentBeliefs and CommunicativeIntent) have several unpaired sets (see Table~\ref{tab:LocalizerSuites} for examples).

To mitigate data contamination and to control for irrelevant surface-level variation in the localizer materials that may affect LMs more than humans (e.g., differences in semantic content of target~vs.~control stimuli), we used GPT-5 (not evaluated in this study) with a 5-shot in-context prompt to generate 100 novel synthetic stimuli for each condition in each suite
(see Appendix~\ref{app:sec:prompts-synthetic-loc-prompts} for all prompts).
To critically assess the quality of the novel synthetic stimuli against the established original stimuli,
we ran a Principal Component Analysis with two components comparing sentence embeddings
from SentenceTransformer’s all-MiniLM-L6-v2 model \citep{reimers-2019-sentence-bert} of the original and synthetic stimulus stories. 
For all conditions except HumanDescr from the LatentBeliefs suite, we confirmed that original and synthetic stimuli did not form clearly separate clusters, indicating that the synthetic stimuli are overall comparable to the original ones. 
For the more divergent HumanDescr condition, the most likely reason is that the original stimuli only described the physical appearance of humans, while the synthetic stimuli covered other aspects as well. This difference, however, does not impact their role as a non-ToM control condition.
More details and visualizations of the PCA results are in Appendix~\ref{sec:app:localizer-pca}.

All localizer items consist of context sentences and end in a task question, which would require a human participant to engage in Theory of Mind reasoning in the target conditions, and
multiple choice answer options. 
More details about all localizer suites are given in Appendix~\ref{app:sec:localizer-suites}.


\subsection{Localization Procedure in LMs}
\label{sec:loc-method}

For an autoregressive transformer model with a hidden dimension $d$,
let $a_{l,t} \in \mathbb{R}^d$ be the output vector of each transformer block of the $l$-th layer at token $t$ of the input sequence (target or control stimulus).
We record the outputs of each unit $i \in \{1, \dots, d \}$, 
which we call \textit{activations}:
$a_{l,t,i}$. 
To represent activations for the entire stimulus, we use activation at the last token before the model is required to produce the answer.
Given a \textit{localizer dataset} with a set of \textit{target stimuli} $S_T$ and a set of \textit{control stimuli} $S_C$, we generate two sets of activations for every unit $(l, i)$ (i.e., target and control activations).
We identify functional subnetworks based on relative differences in activation for $S_T$ in contrast to $S_C$, as measured by an unpaired or paired Welch's $t$-test (the latter used for the simple suites GameBelief and MoralIntent).
We write $\mathcal{T}(i, S_T, S_C)$ for the $t$-test statistic when comparing activations at unit $i$ for target stimuli in $S_T$ and control stimuli in $S_C$.

We use two statistical approaches for localization.
The \textbf{simple} approach defines a metric for unit $i$ under localizer $L$ as $m^L_i = \mathcal{T}\left(i, \bigcup_j S_T^{(j)}, \bigcup_k S_C^{(k)} \right)$, i.e., the relevant $t$-test statistic for the union of all target and control items.
We add a stricter \textbf{conjunctive} approach, which is common in cognitive neuroscience and uses a so-called \textit{minimum statistic} approach \citep{nichols2005valid} to detect significant differences between any pair of relevant target and control items.
This metric for unit $i$ and suite $L$ 
selects the lowest $t$-statistic for all pairwise comparisons: $m^L_i = \min_{j,k}\mathcal{T} \left(i, S_T^{(j)}, S_C^{(k)} \right)$.
For simple suites, the conjunctive approach reduces to the simple.

Based on localizer $L$, we select the \textit{subnetwork} of units $i$, such that $m_i$ is statistically significant at the significance level $\alpha = 0.05$ (modulo appropriate controlling for multiple comparisons).
If that set comprises more than 1\% of all units, we restrict the selection to the top 1\% of units with the highest unique absolute measure $|m_i^L|$. 
For critical ToM localization with each suite $L$, we also construct a control localization condition based on the \textit{least active} subnetwork by selecting the same number of units with the smallest absolute $|m_i^L|$-value from the subset of units whose $m_i^L$ is not significant.
To investigate the causal role of the localized subnetworks through ablation experiments (Section~\ref{sec:ablation-experiments}), units in the subnetwork are set to zero.

Altogether, we consider eight \textit{localizers} 
for identifying relevant subnetworks for subsequent analyses.
We have five \textit{simple localizers}, four of which are obtained from each of the the four localizer suites, and another one from the union of all target and control sets from all localizer suites (the \textbf{all} localizer).
We consider three \textit{conjunctive localizers}, two of which come from the LatentBeliefs and the CommunicativeIntent suites.
A third conjunctive localizer was constructed from the union of these two suites (the \textbf{LB + CI} localizer), as they have non-paired stimuli and were considered to be most relevant for pragmatic reasoning.

\subsection{Models and Evaluation Datasets}
\label{sec:loc-models-datasets}
We run localization experiments on language models from four families: Qwen-2.5, Llama, Falcon-3 and Gemma-2, where both base and fine-tuned models are available (20 models in total).
The models were selected among the models evaluated behaviorally (see Section~\ref{sec:behavioral-eval-models}) to cover a wide range of sizes, while generally performing above chance on ToM and pragmatics.
To assess the role of the localized subnetworks for the performance on these domains, for each of the 20 models we selected 19 to 24 datasets on which the model performed above chance from ToM and pragmatics among the 36 datasets used for behavioral evaluations.



\begin{figure}[t!]
\includegraphics[width=0.48\textwidth]{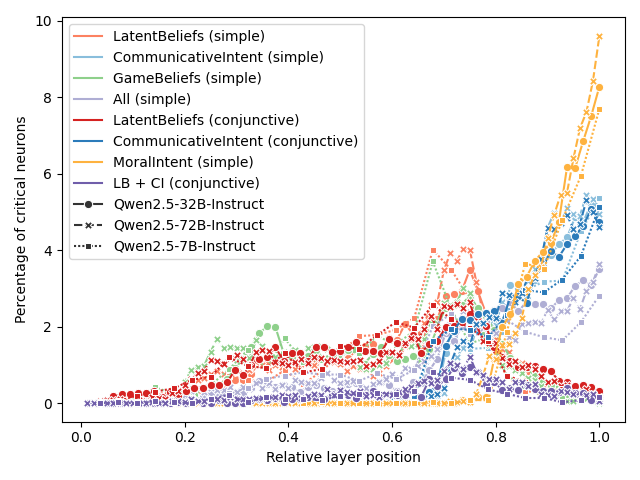}
\caption{Percentage of localized units across model layers (x-axis) identified by different target subnetwork localizers (colors) for Qwen-2.5 Instruct models (shapes).
\label{fig:loc-distribution-qwen}}
\end{figure}

\begin{figure*}[h]
\centering
\includegraphics[width=0.95\textwidth]{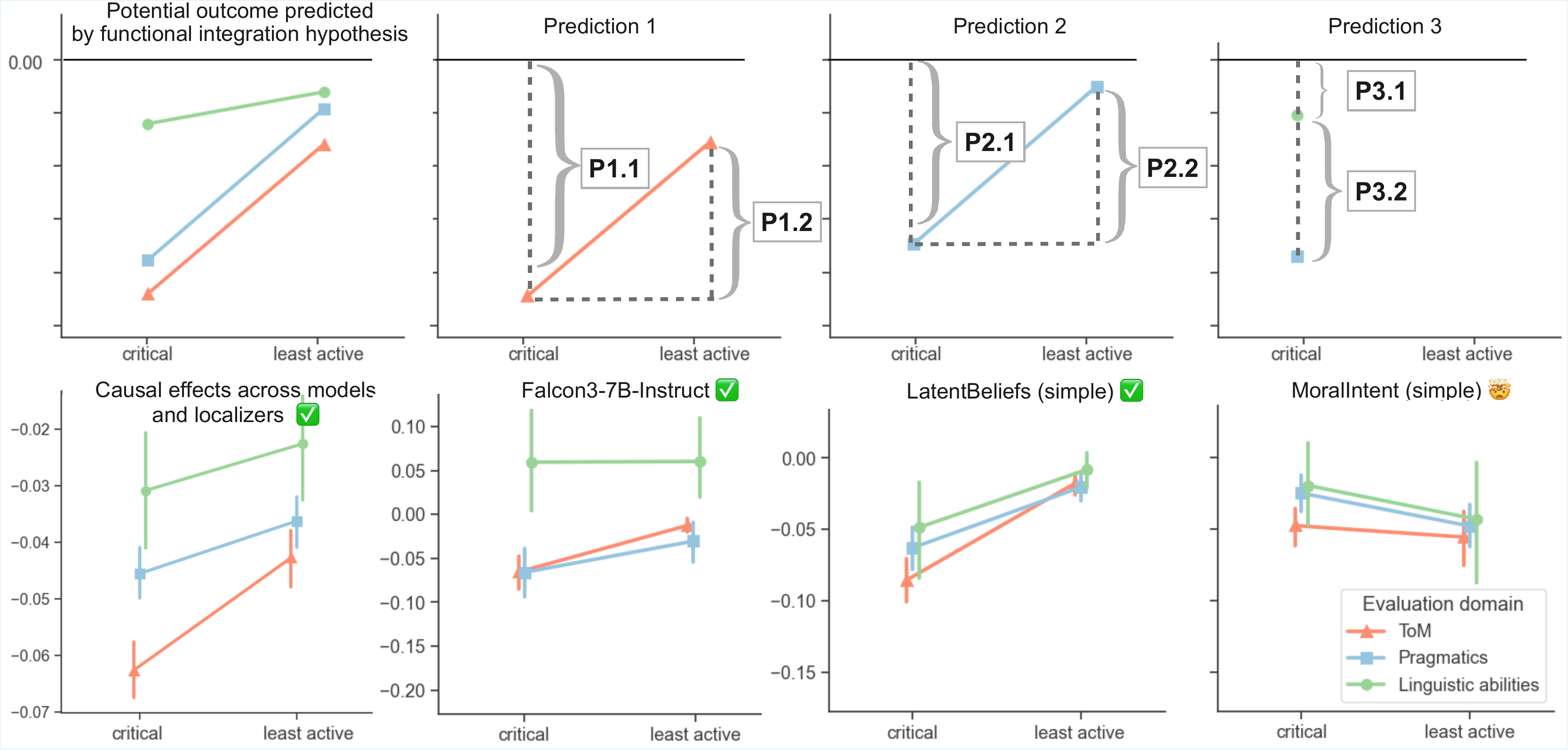}
\caption{
\textbf{All:} Plots show the mean causal effects (y-axis), i.e.,  difference in accuracy relative to the non-ablated baseline when ablating the critical (ToM) subnetworks or the control (least active) sub-networks (x-axis), for different test sets (colors and shapes). 
95\% CIs indicate change in by-dataset accuracy across localizer suites. 
\textbf{Top row:} Fictitious, stylized result that could come from our experiments (left-most plot), and a visualization of the three main predictions made by the functional integration hypothesis.
\textbf{Bottom row:} Four observed outcomes at different levels of analysis. Results from the ``global analysis'' (marginalizing over all language models and localizers) are shown on the left; followed by example results for data from only one language model (averaging over localizers) and two localizer sets (averaging over language models).
Predictions are supported for the ``global analysis,'' but not for all subunits at finer levels of analysis (e.g., the right-most example in the bottom row).
\label{fig:loc-ablation-acc-selected}
}
\end{figure*}

\subsection{Localization Results}
\label{sec:localization-distributions}
The distributions of units selected by the eight localizers are shown in Figure~\ref{fig:loc-distribution-qwen} for three models.
The plot suggests differences in the distributions among localizers; e.g., for CommunicativeIntent highly active units are concentrated in the last layers, while for LatentBeliefs mid to late layers are active. 
Also, the LB + CI localizer seems to be more conservative: often less than 1\% of the models' units were significantly active. 
The patterns appear to be qualitatively similar across model sizes (see Figures~\ref{app:fig:loc-tom-synthetic-critical}--\ref{app:fig:loc-tom-synthetic-random} in the Appendix for full results).

To assess the general robustness of the localizations, we used 10-fold cross-validation, splitting the localizer stimuli into 90:10 localization and testing splits. 
We identified active units on the localization split, recorded the activations of these units for the target and control condition of the testing split, and conducted a two-sided $t$-test ($\alpha=0.05$) on these test activations for each fold, determining whether the units' activity is significantly different. 
Across models, we find the highest generalization rate ($\geq 9$ significant folds) for the localizers all, MoralIntent and GameBeliefs, high rates ($\geq 8$ significant folds) for simple CommunicativeIntent and LatentBeliefs, and lower generalization rates (5--7 significant folds) for conjunctive suites.\footnote{Due to a coding error, results for the LB + CI suite could not be included at the time of writing.}

These results suggest localization is systematic and reproducible, non-random across layers, and that different localizer suites may yield different localized subnetworks, and localizations with the conjunctive suites may be subject to more noise.

\subsection{Ablating Localized Subnetworks}
\label{sec:ablation-experiments}
To assess the causal role of the localized networks, we first ablated, for all 20 models and each  localizer $L$, the ToM network identified by $L$ and the ``least active'' control network for $L$, and then evaluated the performance on selected datasets as described in Section~\ref{sec:behavioral-eval-models} for each ablation.

We test predictions about three causal effects (on ToM, on pragmatics and on baseline general linguistic abilities (BLiMP and SNLI)), which, if observed, conjunctively provide evidence for functional integration (see the upper panel of  Figure~\ref{fig:loc-ablation-acc-selected}):
\begin{enumerate}
        \item[\textbf{P1.1}] ToM-network ablation decreases performance on ToM-related tasks.
        \item[\textbf{P1.2}] CE (causal effect) on ToM-task performance is bigger for ToM ablation than for control.
        \item[\textbf{P2.1}] ToM-network ablation decreases performance on pragmatic tasks.
        \item[\textbf{P2.2}] CE on pragmatic abilities is bigger for ToM ablation than for control ablation.
        \item[\textbf{P3.1}] ToM-network ablation does not decrease performance on baseline linguistic benchmarks.
        \item[\textbf{P3.2}] CE is bigger on pragmatics than on baseline.
\end{enumerate}

These predictions can be tested at different levels of aggregation.
For example, results for some localizers may conform to our predictions better than results from others (see the two right-most plots in the bottom row of Figure~\ref{fig:loc-ablation-acc-selected}).
In light of such variation, the most insightful conclusions are those from a  ``global analysis,'' which marginalizes over all LMs and all localization methods and whose results are shown in the left-most facet in the bottom row of Figure~\ref{fig:loc-ablation-acc-selected}.
In order to test predictions from the ``global analysis'' we inspect the relevant posterior probabilities from a 
Bayesian beta regression model.
\footnote{Model in R syntax: \texttt{accuracy $\sim$ model\_type + model\_size + ablation\_localizer * domain}, where \texttt{ablation\_localizer} is a variable with 17 levels combining the two ablations and eight localizations procedures and the baseline intact condition (results from Section~\ref{sec:results:behavioral-evals}).}
Marginalizing across localizers and models, both P1.1 ($\beta = 0.25 [0.14, 0.35]$) and P1.2 ($\beta=0.06 [0.02, 0.11]$) were credibly borne out in our results, providing evidence for a general causal role of the localized ToM networks for LMs' ToM performance.
Similarly, both P2.1 ($\beta = 0.30 [0.20, 0.39]$) and P2.2 ($\beta=0.07 [0.02, 0.12]$) were credibly borne out, indicating that the ToM subnetworks were generally causally involved in LMs' pragmatic performance.
Finally, the data supported P3.1 (no credible decrease of linguistic benchmark performance under ToM ablation: $\beta=0.13 [-0.10, 0.35]$), but not P3.2 ($\beta=0.17 [-0.07, 0.41]$), offering mild evidence that the causal role of the ToM subnetworks is stronger for pragmatic than general linguistic capabilities, from the ``global'' point of view.
Together, our results support to the idea that ToM and pragmatic reasoning \textit{do} share causally relevant functional units in LMs to a non-trivial degree.

Towards further exploration, we tested predictions separately for each of the eight localizers.
Across models, predictions 1.1, 2.1, and 3.1 were borne out for all localizers (except for CommunicativeIntent (conjunctive) (1.1) and GameBeliefs (3.1)). 
Predictions 1.2 and 2.2 were borne out only for LatentBeliefs (simple \& marginally conjunctive) and GameBeliefs, suggesting that these three localizers mainly drive the results above (shown in Figure~\ref{fig:loc-ablation-acc-acrossModels}).
This suggests that subnetworks tapping into the ToM aspects beliefs, percepts, desires and emotions (captured by these three localizers) might functionally support both ToM and pragmatic performance of LMs.
We note that these three localizers also showed similar distributional patterns in the models (Figure~\ref{fig:loc-distribution-qwen}), different from other localizers.
Results for all localizers and models are shown in the Appendix in Figures~\ref{app:fig:loc-ablation-acc-byLocalizer-part1}--\ref{app:fig:loc-ablation-acc-byLocalizer-part2}, full statistical results are presented in Table~\ref{app:tab:byLoc-hypotheses}.

Moreover, we investigated to which extent the identified subnetworks functionally support mental state related reasoning as opposed to other cognitive skills that may be recruited in ToM and pragmatic tasks.
In addition to the general linguistic ability benchmarks, we evaluated selected models on the entity tracking \citep{kim2023entity} and analogical reasoning tasks \citep{webb2023emergent}. 
When ablating the critical and control subnetworks, e.g., localized by LatentBeliefs and GameBeliefs, we found that the critical ablations had a similar effect on entity tracking performance as on ToM and pragmatic performance across models, but did not as consistently affect other datasets, suggesting that ToM and pragmatic capabilities might indeed also rely on shared entity tracking mechanisms. 
Details of the analysis are reported in Appendix~\ref{app:sec:ablations-g-laoding-datasets}.

For better understanding the impact of the statistical methods for localization, 
we compared the causal effects of ablations of subnetworks localized with the simple~vs.~the conjunctive method for LatentBeliefs (LB) and CommunicativeIntent (CI).
Aggregating causal effects on ToM and pragmatics relative to the baseline (P1.1 and P2.1), we found no credible differences for LB ($\beta=0.05 [-0.03, 0.15]$), nor CI ($\beta=0.06, [-0.03, 0.15]$). 
Relative to the control ablation (P1.2 and P2.2), we also found no credible differences (LB: $\beta=0.10 [-0.03, 0.23]$, CI: $\beta=0.07, [-0.06, 0.20]$). 
The proportion of the units localized by both methods is moderately high (LB = 0.50, CI = 0.66).
This suggests that (i) simple and conjunctive strategies may be interchangeable, and (ii) the observed causal effects may be due to only a subset of the localized subnetworks.

Finally, we compared the causal effects of the ToM ablation between different model sizes using a beta regression model with a three-way interaction.\footnote{Model in R syntax: \texttt{accuracy $\sim$ model\_type + model\_size * ablation\_localizer * domain}.}
Averaged across localizers, we found no credible differences in the causal effect on ToM and pragmatic performance of models of different sizes (L~vs.~M models: ToM: $\beta = 0.15 [-0.12, 0.42]$, pragmatics: $\beta=0.13	[-0.12, 0.38]$; M~vs.~S models: ToM: $\beta = -0.04 [-0.26,	0.18]$, pragmatics: $\beta = -0.11 [-0.41, 0.20]$), suggesting that varying capacity of the models may not directly affect functional integration.

\section{Discussion}
This paper offered five contributions towards addressing the overarching research question of whether LMs contain ``social world models'' --- latent representations of information related to social reasoning about other minds.
(1) We outlined the functional integration hypothesis, according to which internal representations and computational mechanisms are shared between ToM and pragmatic reasoning tasks, and we derived concrete, testable predictions from this view.
(2) We report stringent theory-driven statistical analyses based on data from both behavioral evaluations and causal ablation studies which provide suggestive evidence in favor of the functional integration view.
(3) Improving on previous work \citep{alkhamissi-etal-2025-llm}, 
this paper contributes four novel synthetically generated localization suites with a total of 1400 stimuli based on localizers carefully assembled from prior work in cognitive neuroscience that spans a substantial swath of various aspects of ToM reasoning.
(4) We ported established methods from neuroscience (the conjunctive analysis) into the realm of language models, and presented first results to assess their value.
(5) Finally, we explored a finer-grained analysis of ToM based on the ATOMS framework and found that its subcategories are indeed predictive of benchmark accuracies. 
They also help to interpret the results from the causal ablation studies, where the ablation of localized subnetworks addressing those ATOMS categories that were part of the best explanations of the behavioral data 
also had the strongest causal effect of interest on LMs' ToM and pragmatic capabilities, 
although it remains puzzling why ablating the subnetwork identified by the CommunicativeIntent localizer did \textit{not} have the predicted causal effect.

Even though we found support for various causal effects predicted by a functional integration view, our results hinge on whether the datasets used for localizations and behavioral analysis indeed operationalize the right kinds of concepts (general ToM, specific aspects in the ATOMS framework, pragmatic reasoning etc.). 
As is the case with human experiments, the validity of methods and operations require converging evidence from many like-minded investigations and 
replications.
What is more, while it is commonly assumed that ToM and pragmatics tasks recruit reasoning about mental causes, we should be open to the possibility that, also in humans, explicit social reasoning trades off with cheaper heuristics \cite{Rubio-Fernandez2024:Cultural-evolut}. 
Additionally, both tasks may share not only social reasoning, but likely also other general intelligence capabilities. 
While the exact nature of such general mechanisms involved in ToM and pragmatics remains elusive (for both humans and machines), our novel large localizer datasets and fine-grained statistical methodology for establishing functional integration between such tasks show one potential path towards contributing to this deep issue. 

Together, these considerations underline the importance of an interdisciplinary perspective on language abilities, closely comparing human and machine solutions to superficially similar tasks. 
In this spirit, the present paper contributes by investigating LMs' functional mechanisms of different aspects of social reasoning at a high resolution, offering exciting avenues for future investigations of the exact relation between different capabilities and of the architectural pressures shaping LMs' ``social world models.''

\section*{Limitations}
The presented functional localizations and evaluations make a number of design choices in the procedure which are known to potentially impact the results \citep{bommasani2023holistic}, and future work should evaluate the robustness of the results to such design choices.

For the baseline and ablation behavioral evaluations, we chose to use means over the conditional log probabilities of the answer tokens.
For the subnetwork localization, we used one general instruction and one specific prompt format across stimuli.
All localizer stimuli contained a forced-choice task, which was formulated as a sentence continuation task with two answer options for all localizer suites, except for CommunicativeIntent, for which the task was a meta-linguistic judgment task to select whether an utterance was literal, ironic, deceitful or meaningless.

Furthermore, to ablate the localized subnetwork, the identified units were set to zero; however, alternative approaches also exist \citep{heimersheim2024use}.
The size of the ablated subnetworks is also set to the minimum of the top-most active 1\% of the units and all significantly active units; however, smaller or larger ablation rates could also be explored \citep[cf.][]{alkhamissi-etal-2025-llm}.

The ATOMS annotations used for more fine-grained analyses of ToM were, in part, provided by the authors, resulting in a low number of annotations and potential noise due to ambiguity of commonly used ToM and pragmatics evaluation benchmarks. 
The annotations could be improved upon in future work through, e.g., crowdsourcing.

Furthermore, future work could expand the selection of datasets used to evaluate the general linguistic abilities of models beyond the BLiMP and SNLI datasets. 
Additionally, all datasets focus exclusively on English, a resource-rich language; in particular, pragmatics datasets in other languages may differ with respect to the cultural and social norms that must be reasoned about.
The precision and diversity of the pragmatics datasets may also vary between the selected benchmarks, e.g., due to semi-automated construction of some of the materials.
The ToM datasets contain a large proportion of false-belief tasks, whereas real-world language use may provide more evidence for true-belief-based reasoning, which may therefore be strongly represented in LMs' training data.
Finally, since some of our datasets have been publicly available for several years, there is a risk that the training data of recent models are contaminated with these datasets.

Although the synthetic generation of novel localizer materials was based on validated materials from neuroscience research and allowed studying language models at a larger scale than related previous work, these new materials have not been validated in human fMRI studies.
Therefore, while drawing inspiration from cognitive neuroscience research, our conclusions cannot be directly translated to claims about human-likeness of the functional organization of LMs.

In general, localized subnetwork ablation experiments offer only limited, coarse-grained insights about LMs' internal organization; achieving a more precise understanding of the causal role of single units for ToM and pragmatic performance will require alternative techniques like causal tracing or mediation methods, such as activation and attribution patching and circuit discovery \citep{wang2022interpretability}.


\section*{Acknowledgments}
JFK, AM and MF were supported by the Volkswagen Foundation through a Momentum grant. 
MF is a member of the Machine Learning Cluster of Excellence at University of T\"ubingen, EXC number 2064/2 – Project number 39072764.
PT and MF gratefully acknowledge support by the state of Baden-
W\"urttemberg through bwHPC and the German Research Foundation
(DFG) through grant INST 35/1597-1 FUGG, and the data storage service SDS@hd supported by the Ministry of Science, Research and the Arts Baden-W\"urttemberg (MWK) and the German Research Foundation (DFG) through grant INST 35/1803-1 FUGG and INST 35/1804-1 LAGG.
\bibliography{custom}

\appendix

\section{Behavioral Evaluation}
\label{sec:app:behavioral-eval}
\label{sec:app:behavioral-eval-datasets}


\subsection{ATOMS Categories}
\label{sec:app:atoms}
The ATOMS framework \citep{beaudoin2020systematic} is an approach to classify ToM experiments based on the specific capabilities they investigate. While \citet{beaudoin2020systematic} use the framework to mostly classify ToM measures into disjoint categories (such that each task is placed in the best-fitting ATOMS category only), in practice, many experiments address more than one aspect of ToM. We follow \citet{ma-etal-2023-towards-holistic} in that we characterize ToM tasks according to all ATOMS aspects for which we deem them relevant. 

The ATOMS framework encompasses seven broad categories, each with subdivisions: 1) The \textbf{beliefs} category targets the capability to ascribe beliefs to actors that may be different from one’s own beliefs, such as required in the Sally-Anne task \citep{baron1985does}. This also includes second-order beliefs, i.e., beliefs about other people’s beliefs. 2) The \textbf{intentions} category targets the capability to ascribe intentions to people and to distinguish these intentions from their actions. 3) The \textbf{desires} category encompasses tasks that require ascribing desires to actors. 4) \textbf{Emotions} refers to the ability to predict other people’s emotions from events, and to understand that other people may hide or regulate their emotions such that they are not directly accessible. 5) The \textbf{knowledge} category concerns being aware of different people having different knowledge which may result in different actions. 6) \textbf{Percepts} 
include perspective taking (visual and otherwise), and being aware that actors' actions may differ depending on what they have perceived. 7) The \textbf{mentalistic understanding of non-literal communication} category includes tasks involving irony, humor, or lies.
For detailed introductions of the categories, see \citet{beaudoin2020systematic, ma-etal-2023-towards-holistic}.

The ATOMS categories (on a coarse-grained level) addressed by the test datasets used in this study are listed in Table~\ref{tab:datasets-atoms}.

\begin{table*}
\footnotesize
    \begin{tabular}{c|c|c|c|c|c|c|c}
         Dataset & beliefs & intentions & desires & emotions & knowledge & percepts & NLC \\
         BigToM \citep{gandhi2023understanding} & & & & & & & \\
         $\bullet\text{ forward}$ & \checkmark & $\times$ & $\times$ & $\times$ & $\times$ & \checkmark & $\times$ \\
         $\bullet\text{ backward}$ & \checkmark & $\times$ & $\times$ & $\times$ & $\times$ & \checkmark & $\times$ \\
         EmoBench  \citep{sabour-etal-2024-emobench} & \checkmark & $\times$ & $\times$ & \checkmark & $\times$ & $\times$ & $\times$ \\
         EPITOME \citep{jones-etal-2024-comparing-humans} & & & & & & & \\
         $\bullet\text{ false belief}$ & \checkmark & $\times$ & $\times$ & $\times$ & $\times$ & $\times$ & $\times$ \\
         $\bullet\text{ recursive mindreading}$ & \checkmark & $\times$ & $\times$ & $\times$ & $\times$ & $\times$ & $\times$ \\
         EWoK \citep{ivanova2025elements} & & & & & & & \\
         $\bullet\text{ agent properties}$ & \checkmark & $\times$ & $\times$ & $\times$ & \checkmark & \checkmark & $\times$ \\
         $\bullet\text{ social properties}$ & $\times$ & $\times$ & $\times$ & \checkmark & $\times$ & $\times$ & $\times$ \\
         $\bullet\text{ social interactions}$ & $\times$ & \checkmark & $\times$ & \checkmark & $\times$ & $\times$ & $\times$ \\
         FauxPasEAI \citep{shapira-etal-2023-well} & & & & & & & \\
         $\bullet\text{ Q1 \& Q4} $& \checkmark & \checkmark & $\times$ & $\times$ & $\times$ & $\times$ & \checkmark \\
         OpenToM \citep{xu-etal-2024-opentom} & & & & & & & \\
         $\bullet\text{ coarse-grained first-order} $& \checkmark & $\times$ & $\times$ & $\times$ & $\times$ & \checkmark & $\times$ \\
         $\bullet\text{ coarse-grained second-order}$ & \checkmark & $\times$ & $\times$ & $\times$ & $\times$ & \checkmark & $\times$ \\
         $\bullet\text{ fine-grained first-order}$ & \checkmark & $\times$ & $\times$ & $\times$ & $\times$ & \checkmark & $\times$ \\
         $\bullet\text{ fine-grained second-order}$ & \checkmark & $\times$ & $\times$ & $\times$ & $\times$ & \checkmark & $\times$ \\
         $\bullet\text{ multi-hop first-order}$ & \checkmark & \checkmark & \checkmark & \checkmark & $\times$ & \checkmark & $\times$ \\
         $\bullet\text{ multi-hop second-order} $& \checkmark & \checkmark & \checkmark & \checkmark & $\times$ & \checkmark & $\times$ \\
         $\bullet\text{ attitude}$ & \checkmark & $\times$ & \checkmark & \checkmark & $\times$ & \checkmark & $\times$ \\
         SimpleToM \citep{gu2024simpletom} & $\times$ & \checkmark & $\times$ & $\times$ & $\times$ & $\times$ & $\times$ \\
         Social IQa \citep{sap-etal-2019-social} & $\times$ & \checkmark & $\times$ & \checkmark & $\times$ & $\times$ & $\times$ \\
         ToMi \citep{le-etal-2019-revisiting} & & & & & & & \\
         $\bullet\text{ first-order false-belief}$ & \checkmark & $\times$ & $\times$ & $\times$ & $\times$ & $\times$ & $\times$ \\
         Triangle-COPA \citep{maslan2015one} & $\times$ & \checkmark & $\times$ & \checkmark & $\times$ & $\times$ & $\times$ \\
    \end{tabular}
    \caption{Classification of ToM datasets according to the ATOMS framework \citep{beaudoin2020systematic}. NLC stands for “non-literal communication”. This table takes inspiration and, in part, annotations from those in \citet{ma-etal-2023-towards-holistic} and \citet{saritacs2025systematic}, and includes our own annotations of additional datasets and partial revisions of the annotations. 
    }
    \label{tab:datasets-atoms}
\end{table*}

\subsection{Datasets}
\label{sec:app:datasets}

\subsubsection{Theory-of-Mind Datasets}

\textbf{BigToM} \citep{gandhi2023understanding} is a benchmark on social reasoning / ToM capabilities. The individual items in the dataset are paired false-belief and true-belief situations generated by an LLM in accordance to a predefined framework. The questions in BigToM involve either forward ToM, i.e., inferring a person's beliefs based on what they did or did not observe, and backward ToM, i.e., inferring a person's beliefs based on their subsequent actions. 
The items contain two answer options per item.
We consider an LM's prediction for a single item correct only if the correct answer options are selected both for the false-belief and the true-belief situations.

\textbf{EmoBench} \citep{sabour-etal-2024-emobench} is a benchmark on emotional intelligence assessing both emotional understanding and its application. The EmoBench tasks focus on aspects such as complex emotions, perspective taking, emotional cues, personal beliefs and experiences, and subcategories of these aspects. The associated questions are phrased as multiple choice questions, where we compare the correct answer with all incorrect answers and consider a model's choice as correct if the conditional probability of the correct answer, given then context, is higher than all conditional probabilities of the incorrect answers. 

\textbf{EPITOME} \citep{jones-etal-2024-comparing-humans} is a battery of six ToM-related experiments, of which we use two (in addition to the scalar implicature task described in the following section): recursive mindreading and false belief. The recursive mindreading task tests the representation of recursively embedded mental states. The false belief task uses a setting similar to the Sally-Anne task \citep{baron1985does}. All questions are phrased as multiple-choice questions.

\textbf{EWoK} \citep{ivanova2025elements} is a framework for evaluating world modeling abilities. It consists of eleven tasks, out of which we use three: agent properties, social properties, and social interactions. The agent properties split requires to estimate which one of two propositions about the properties of an agent (such as mental states or preferences) is true, given some context. In the social properties split, the two alternatives are socially relevant agent properties (e.g., character traits). In the social interactions split, there are two agents, and the answer alternatives are possible social relations between them (e.g., cooperating or competing with each other).
Each item contains two contexts and two answer options, each being the correct answer for one of the contexts.
We apply the same procedure for retrieving predictions for each item as for BigToM. 

\textbf{FauxPasEAI} \citep{shapira-etal-2023-well} is a dataset intended for evaluating the ability to recognize situationally inappropriate utterances. The original dataset has four questions Q1-Q4, of which we exclude Q2 because it is an open-ended question and Q3 because it does not directly concern ToM abilities. The remaining questions Q1 and Q4 are both “yes/no”-questions, where Q1 targets recognizing whether there is a situationally inappropriate utterance, while Q4 concerns the knowledge available to the person uttering it.

\textbf{OpenToM} \citep{xu-etal-2024-opentom} is a ToM benchmark for LLM evaluation consisting of several ToM-related subtasks where an object is moved in the presence or absence of observers (following the schema of the Sally-Anne test, \citeauthor{baron1985does}, \citeyear{baron1985does}). The dataset encompasses 1) first-order false belief tasks with both coarse-grained questions (which ask about somebody’s belief whether the location has changed) and fine-grained questions (which ask about somebody’s belief about the exact location), 2) second-order false belief tasks, 3) multi-hop tasks, which are modifications of the false-belief tasks aiming to address the understanding of social norms, and 4) an attitude task that concerns the attitude of an observer towards the mover’s action.

\textbf{SimpleToM} \citep{gu2024simpletom} is a dataset designed to test LLMs' abilities to infer mental states implicitly. This is tested using three types of questions, namely, mental state questions, behavior prediction questions, and judgement questions. We only use the behavior prediction split, where the task is to predict which of two actions a person is more likely to perform next in some context (based on implicit reasoning about their mental state).

The \textbf{Social IQa} dataset \citep{sap-etal-2019-social} is designed to test language models' competence in the areas of social and emotional intelligence. It consists of descriptions of situations and multiple-choice questions about aspects such as motivations, reactions, and social consequences.

\textbf{ToMi} \citep{le-etal-2019-revisiting} is a synthetically constructed dataset of situations resembling the Sally-Anne test \citep{baron1985does}, i.e., where a false-belief is created by an object being moved in the absence of a person. The dataset encompasses questions about first- and second-order true and false beliefs as well as “reality” and “memory” questions. We only use first-order false-beliefs from \citep{sap-etal-2022-neural}, as is also done by \citep{alkhamissi-etal-2025-llm}.
For this dataset, we additionally use the following instruction and format following \citet{alkhamissi-etal-2025-llm}: ``The following multiple choice questions is based on the following story. The question is related to Theory-of-Mind. Read the story and then answer the questions. Choose the best answer from the options provided by printing it as is without any modifications. \textbackslash n Story: \{story\} \textbackslash n Question: \{question\} \textbackslash n Options: \{answer options\} \textbackslash n Answer:''.

The \textbf{Triangle-COPA} dataset \citep{maslan2015one} is a commonsense psychology dataset measuring the ability to infer mental or emotional causes and social relations from actions. The agents are represented as geometric shapes (e.g., triangles), which perform some action. A following binary choice question then offers two motivations for the observed behavior, and the task is to identify the more plausible of them.

\subsubsection{Pragmatics Datasets}

From \textbf{EPITOME} \citep{jones-etal-2024-comparing-humans} (see also the section above), we include the scalar implicature task among the pragmatics experiments. This task tests whether \textit{some} is more likely to be interpreted as meaning \textit{not all} when the speaker has complete knowledge (because then they could have said \textit{all} as well).

The dataset from \citet{hu-etal-2023-fine}, here called \textbf{HuFloyd}, consists of short stories involving non-literal utterances, and questions about the actual intention of the person uttering them. For each story and utterance, the answer options include the target intention, the literal interpretation, and non-literal distractors (except for the coherence task, where there are only two answer options). The dataset targets a range of pragmatic phenomena, namely, deceits, indirect speech, irony, maxims, metaphor, humour, and coherence.

\textbf{IMPPRES} \citep{jeretic-etal-2020-natural} consists of an implicature and a presupposition task, both of which are also part of the PUB benchmark, through which we access this dataset \citep[see below]{sravanthi-etal-2024-pub}. In the implicature task, a premise and a hypothesis are given which involve a scalar implicature trigger. The task is then to judge whether the hypothesis is true given the premise or not. The presupposition task is structured analogously, but involves presupposition triggers. We modify the questions and answer options relative to PUB such that instead of asking whether the hypothesis is true or not true, they ask whether it follows from the premise (such that the answer options are: ``The hypothesis contradicts the premise.'', ``The hypothesis follows from the premise.'', ``The hypothesis is in no clear relationship with the premise.'').

The dataset called \textbf{LUDWIG} \citep{george2020conversational} assesses the ability to interpret conversational implicature correctly. Each stimulus consists of a short dialogue, where a question is followed by a response that does not directly entail “yes” or “no”, but from which the answer can be inferred via conversational implicature. The task then is to judge whether the response means “yes” or “no”.

\textbf{PUB} \citep{sravanthi-etal-2024-pub} is a large benchmark of 14 pragmatics-related tasks for LLM evaluation. Of these tasks, apart from the two ImpPres tasks described above, we use the following four: 1) indirectness classification (i.e., distinguishing direct from indirect responses), 2) agreement detection (i.e., assessing whether two speakers agree), 3) understanding sarcasm, 4) deictic question answering (i.e., correctly inferring whether an object is in some location based on deictic expressions in a dialogue). All these tasks are phrased such that there are exactly two answer options. For agreement detection, we modify the answer prefix to be ``Speaker 2 '', and the answer options to be ``agrees'' and ``disagrees''. For indirectness classification, we modify the answer prefix to be ``The response is '', and the answer options to be ``direct'' and ``indirect''.


\textbf{SarcV2} \citep{oraby-etal-2016-creating} is a corpus of utterances from social media data, which are to be classified into sarcastic and not-sarcastic utterances. The corpus is divided into three subcorpora (generic, hyperbole, and rhetorical questions). We use all three subcorpora without making more fine-grained distinctions.

\subsubsection{Other Datasets}

For comparison, we use items from two datasets testing LMs' general linguistic performance.
First, across behavioral evaluations and functional localization, we use \textbf{BLiMP} \citep{warstadt-etal-2020-blimp-benchmark}, a widely known benchmark on grammatical phenomena in English. BLiMP consists of minimal pairs, where in each pair, one sentence is grammatical and one is ungrammatical. LMs are evaluated by comparing the probabilities they assign to the grammatical and ungrammatical sentences. We use a subset of 10 sample items per category.
Second, for behavioral evaluations we also use \textbf{SNLI} \citep{bowman2015large}, a natural language inference benchmark in English. 
It consists of pairs of sentences, a premise and a hypothesis. The task is to identify whether the relation between these two sentences is entailment, contradiction or in neutral relation. 
The models are evaluated by comparing the conditional probabilities the LM assigns to the labels (e.g., ``The hypothesis follows from the premise.'').
The label to which the model assigns the highest probability given the sentence pair is the model's prediction. 
We use a subset of 670 items, to match the size of the BLiMP dataset.

\subsection{Prompt Formatting}
\label{sec:app:eval-dataset-formatting}
The following datasets contain and answer options which can be considered simple continuations of the statement or partial sentence posing the target task: EPITOME, EWoK, SimpleToM, BigToM, HuFloyd, and LUDWIG.
When models that have chat templates were evaluated on these datasets, the scored answer options were formatted as part of the user turn content.

For the remaining datasets, the optional task, the context and the question or trigger statement were part of the user content, while the answer options were scored as the assistant turn content.

No additional task instructions or prompt formatting were used for any datasets beyond the instructions or formats provided in the original sources of the datasets or described above. 
Only for the following datasets the answer prefix ``The answer is: '' was used: FauxPasEAI, SocialIQa, EmoBench.

\subsection{Models}
\label{sec:app:behavioral-eval-models}

The following models from seven families were evaluated in the behavioral analysis. All models' weights were accessed through the HuggingFace platform. The HuggingFace model names are listed below.

\noindent \textbf{OLMo}: allenai/OLMo-1B-0724-hf, allenai/OLMo-7B-0724-hf, allenai/OLMo-7B-0724-Instruct-hf
 
\noindent \textbf{Pythia}: EleutherAI/pythia-1b-deduped, EleutherAI/pythia-1.4b-deduped, EleutherAI/pythia-2.8b-deduped, EleutherAI/pythia-6.9b-deduped, EleutherAI/pythia-12b-deduped 

\noindent \textbf{Mistral}: mistralai/Mistral-7B-Instruct-v0.3

\noindent \textbf{Llama-2}: meta-llama/Llama-2-7b-hf, meta-llama/Llama-2-7b-chat-hf, meta-llama/Llama-2-13b-hf, meta-llama/Llama-2-13b-chat-hf, meta-llama/Llama-2-70b-hf, meta-llama/Llama-2-70b-chat-hf

\noindent  \textbf{Llama-3}: meta-llama/Llama-3.2-1B, meta-llama/Llama-3.1-8B, meta-llama/Llama-3.1-8B-Instruct, meta-llama/Llama-3.1-70B, meta-llama/Llama-3.3-70B-Instruct

\noindent \textbf{Falcon}: tiiuae/Falcon3-1B-Base, tiiuae/Falcon3-1B-Instruct, tiiuae/Falcon3-3B-Base, tiiuae/Falcon3-3B-Instruct, tiiuae/Falcon3-7B-Base, tiiuae/Falcon3-7B-Instruct, tiiuae/Falcon3-10B-Base, tiiuae/Falcon3-10B-Instruct

\noindent \textbf{Gemma}: google/gemma-2-2b, google/gemma-2-9b, google/gemma-2-9b-it, google/gemma-2-27b, google/gemma-2-27b-it

\noindent \textbf{Qwen}: Qwen/Qwen2.5-0.5B, Qwen/Qwen2.5-0.5B-Instruct, Qwen/Qwen2.5-1.5B, Qwen/Qwen2.5-1.5B-Instruct, Qwen/Qwen2.5-3B, Qwen/Qwen2.5-3B-Instruct, Qwen/Qwen2.5-7B, Qwen/Qwen2.5-7B-Instruct, Qwen/Qwen2.5-14B, Qwen/Qwen2.5-14B-Instruct, Qwen/Qwen2.5-32B, Qwen/Qwen2.5-32B-Instruct, Qwen/Qwen2.5-72B, Qwen/Qwen2.5-72B-Instruct

\subsection{Statistical Analyses}
\label{sec:app:behavioral-eval-stats}
Below, we provide details on how the predictions P1--P3 in Section~\ref{sec:behavioral-eval} are operationalized and tested statistically.
For P2, we used a Bayesian Beta regression model, which predicts the average accuracy of an LM on a given dataset (dependent variable, bound between 0 and 1) according to a Beta distribution, where the predictor for the mean of the distribution is parameterized by several independent variables. 
For each of the independent variables (i.e., intercept, model family, model size, model training type, dataset type, dataset domain) respective regression coefficients are estimated. 
For P2, we hypothesized the following: if computational mechanisms are shared between pragmatics and ToM, there will be no credible differences in the accuracy on ToM~vs.~pragmatics tasks, when controlling for other factors which may lead to differences independently of these mechanisms (e.g., model size), which will be borne out statistically as a non-credible regression coefficient for the predictor ``domain'' (ToM or pragmatics), which we report in Section~\ref{sec:results:behavioral-evals}.

For P3, we used two versions of a Bayesian Beta regression model predicting the independent variable ``accuracy on pragmatics`` (for each dataset for each model) with the dependent variables model size, model type, and either the average accuracy on linguistic benchmarks BLiMP and SNLI (model M0), or on ToM (model M1). 
As standard in leave-on-out cross-validation, we calculated the expected log predictive density (ELPD) of each of the two models, by iteratively fitting each model on the empirical data but a left out data point (done through the R package \texttt{brms}, \citealp[]{burkner2017brms}).
We then compared the two models by evaluating the log predictive density of the held-out observation under each model, computing a $p$-value for this using Lambert’s $z$-score method. 

\subsection{Additional Results}
\label{sec:app:behavioral-eval-analyses}
\begin{figure}
\includegraphics[width=0.48\textwidth]{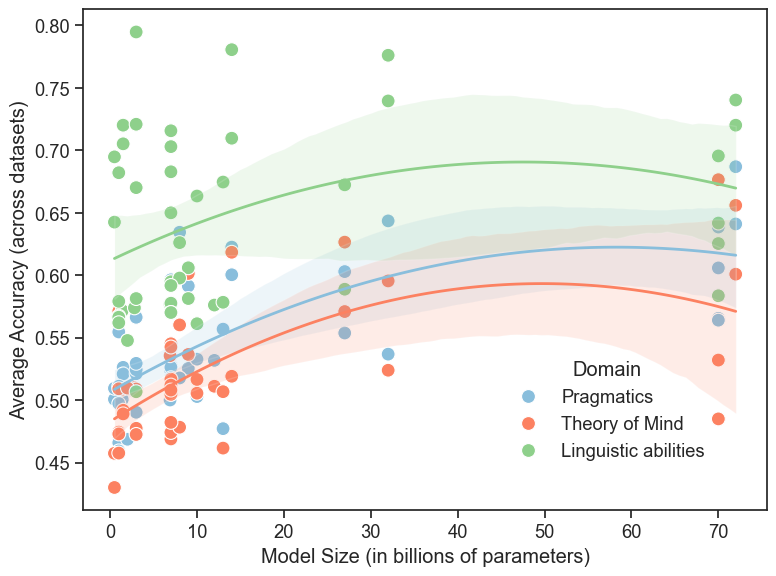}
\caption{Average accuracy (y-axis) for the domains pragmatics, theory of mind and general linguistic abilities (color) by model size (x-axis).\label{app:fig:acc-syntax-size}}
\end{figure}

We show the models' average accuracies on the linguistic benchmarks BLiMP and SNLI, next to the accuracies on ToM and pragmatics in Figure~\ref{app:fig:acc-syntax-size}. 
Consistent with the literature, accuracy on the complex domains ToM and pragmatics increases over model size. However, just model size or model type (base or fine-tuned) are not sufficient for explaining away the connection between pragmatic and ToM performance of LMs, as shown via model comparison in Section~\ref{sec:results:behavioral-evals}.

\begin{figure*}
\includegraphics[width=\textwidth]{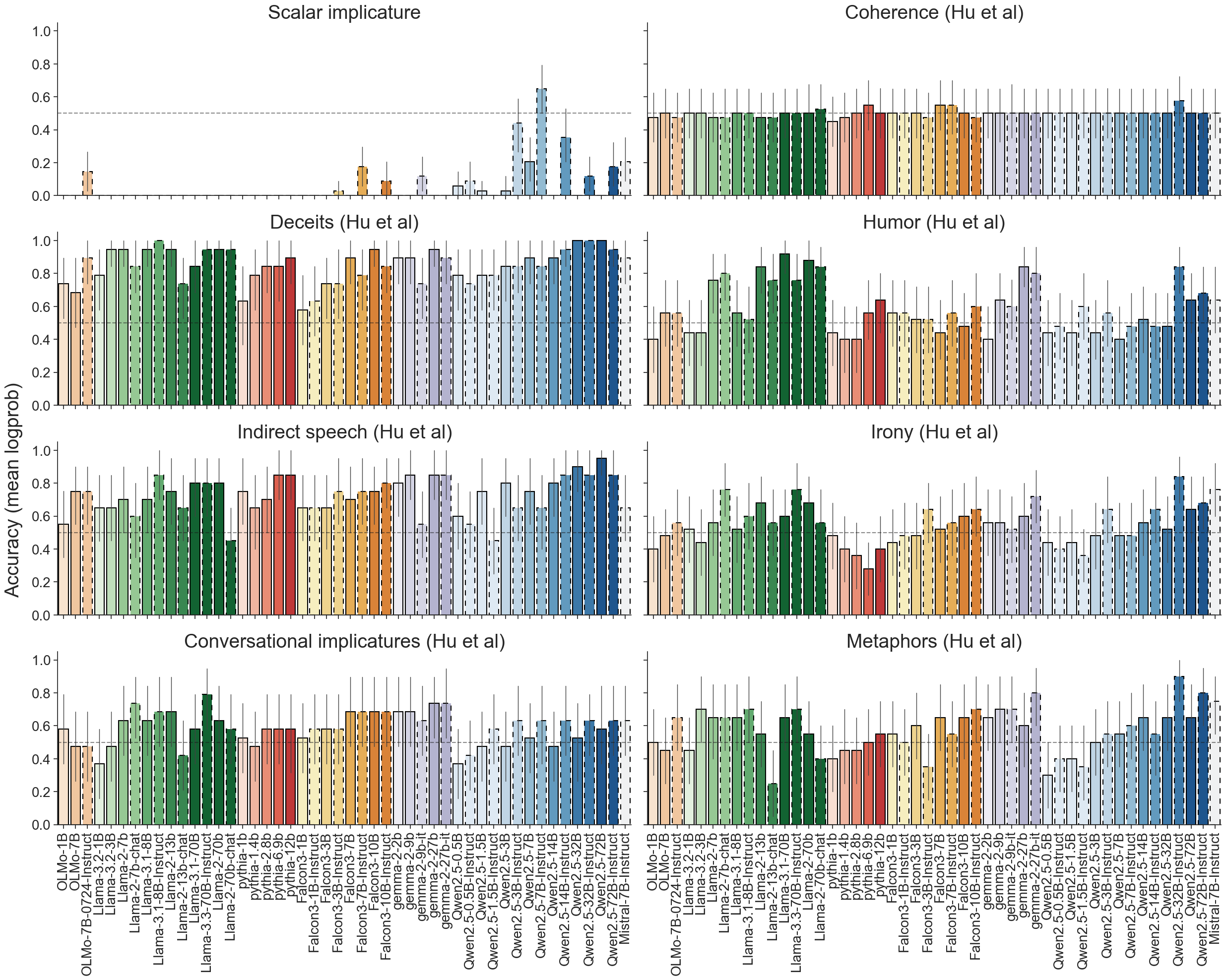}
\caption{Detailed overview of the accuracy (y-axis) results of all models (x-axis) on the first eight pragmatic datasets (facets).\label{app:fig:prag-acc-part1}}
\end{figure*}
\begin{figure*}
\includegraphics[width=\textwidth]{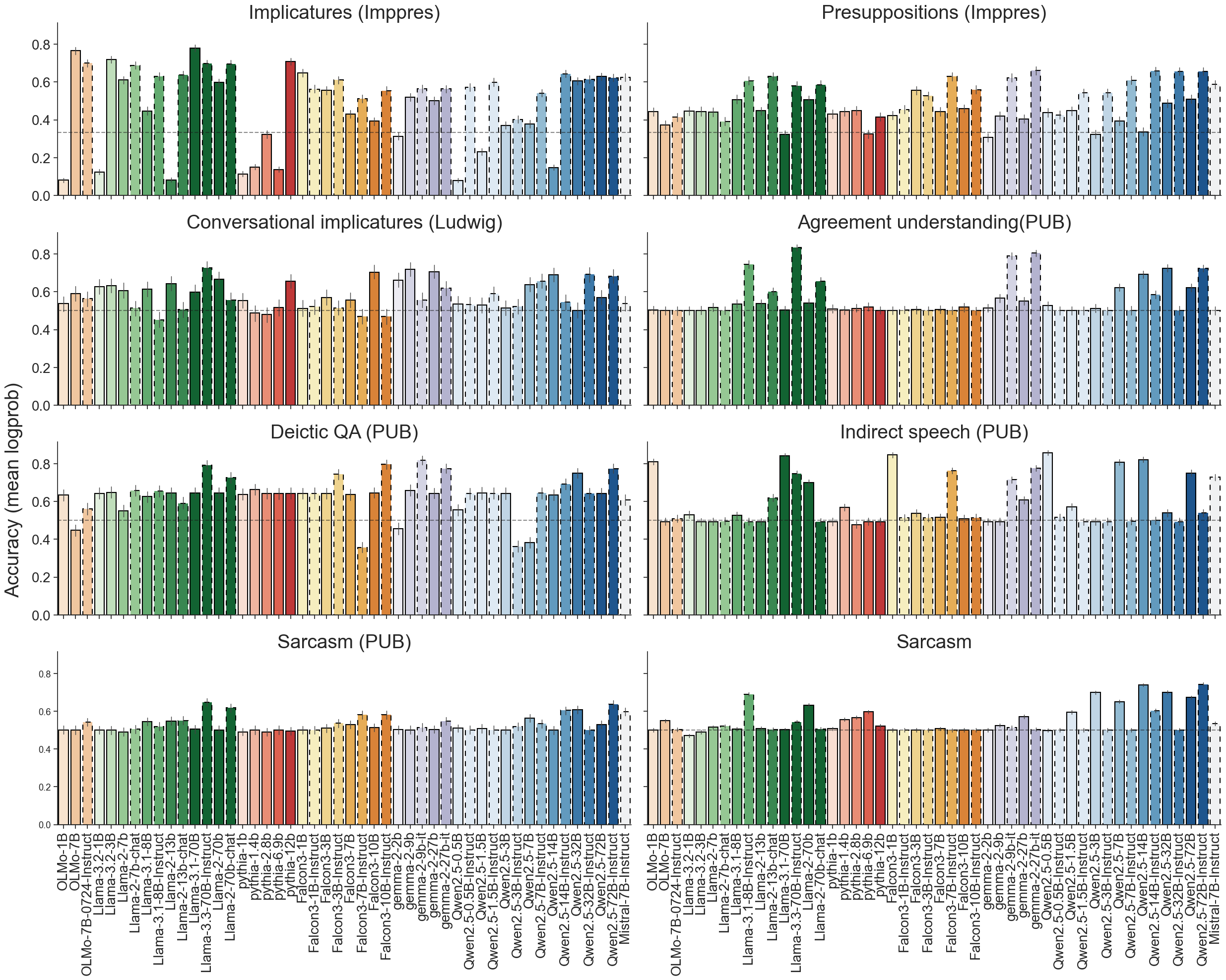}
\caption{Detailed overview of the accuracy (y-axis) results of all models (x-axis) on the remaining eight pragmatic datasets (facets).\label{app:fig:prag-acc-part2}}
\end{figure*}

\begin{figure*}
\includegraphics[width=\textwidth]{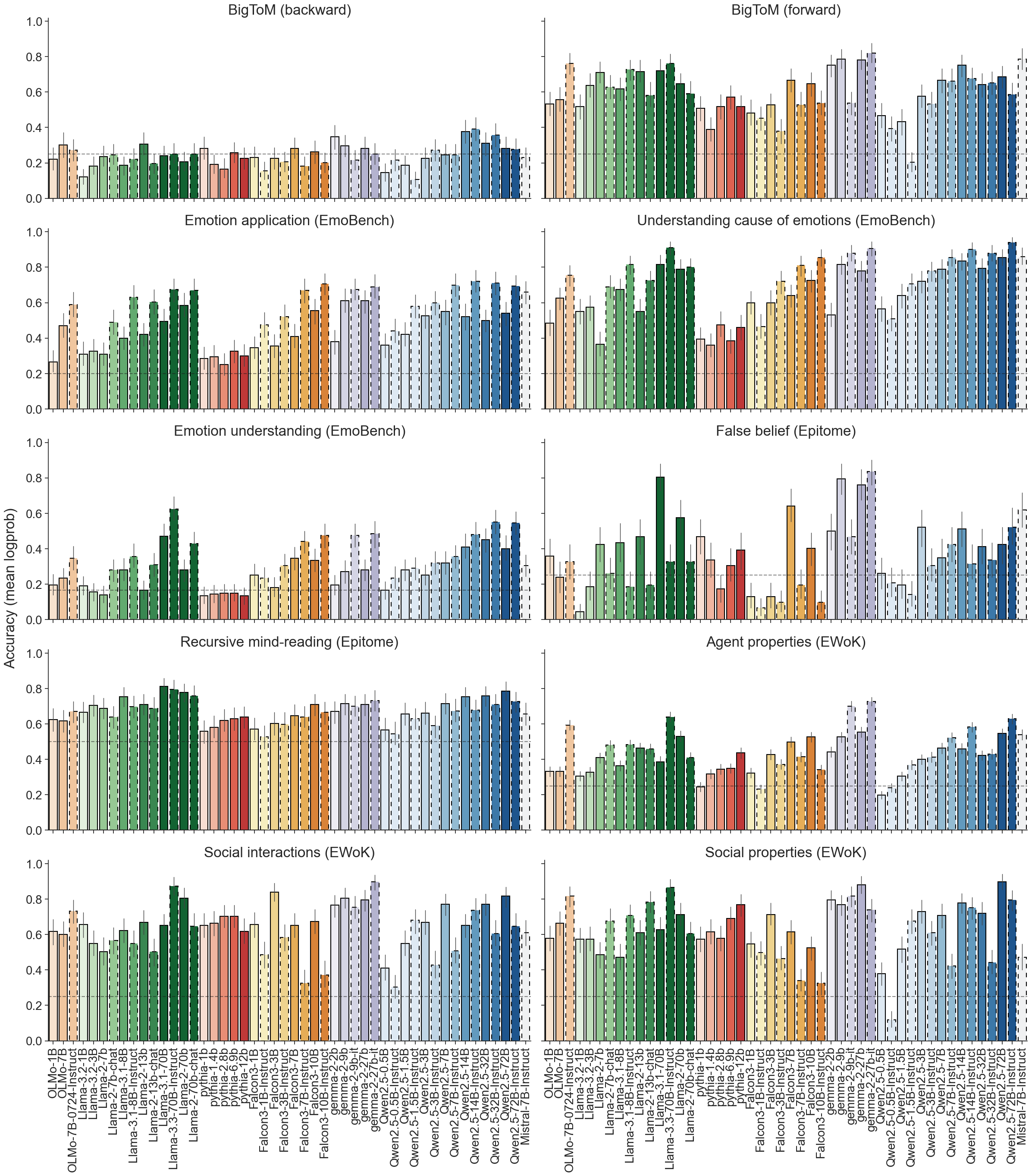}
\caption{Detailed overview of the accuracy (y-axis) results of all models (x-axis) on ten of the ToM datasets (facets).\label{app:fig:tom-acc-part1}}
\end{figure*}
\begin{figure*}
\includegraphics[width=\textwidth]{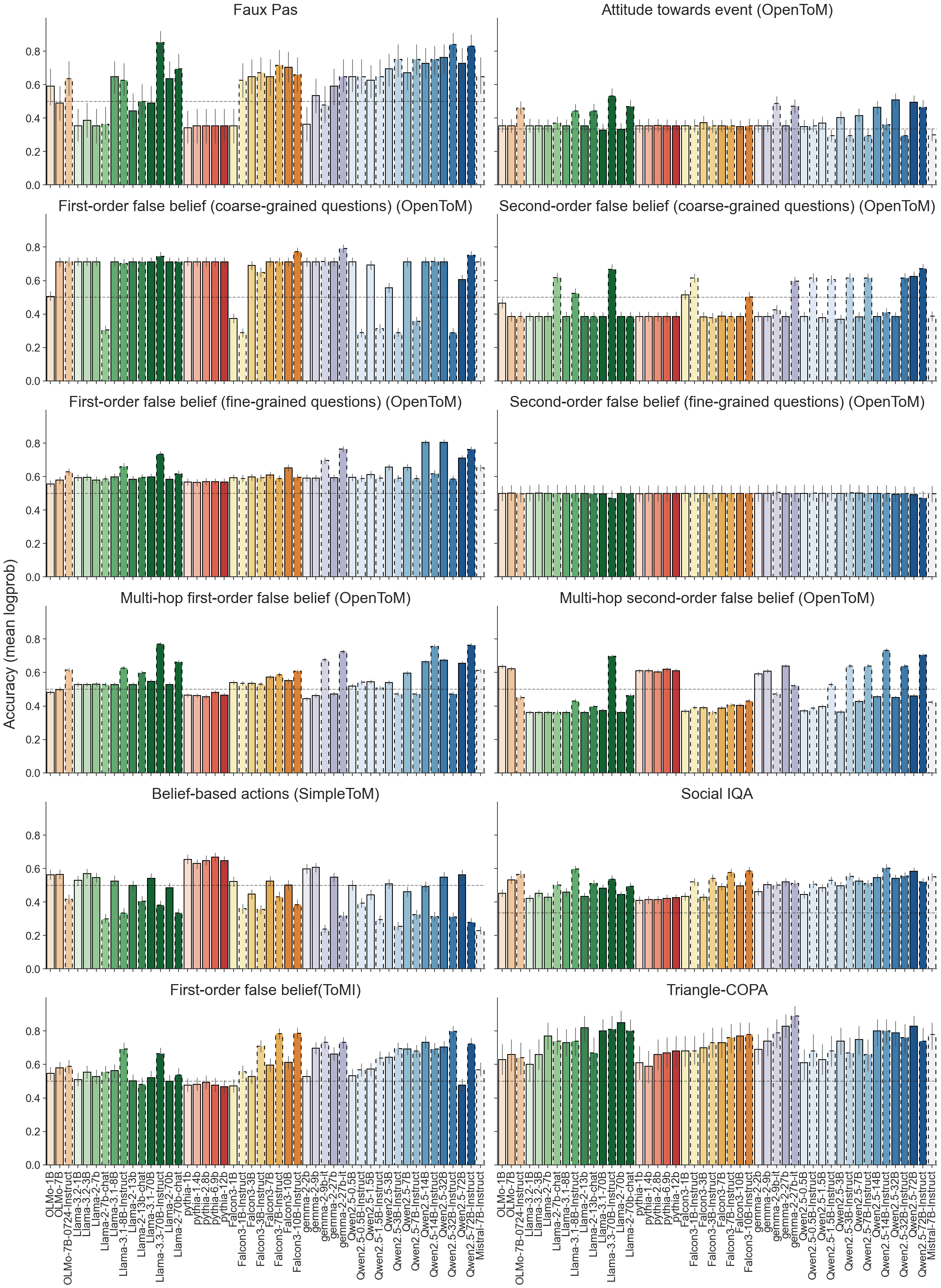}
\caption{Detailed overview of the accuracy (y-axis) results of all models (x-axis) on the remaining twelve ToM datasets (facets).\label{app:fig:tom-acc-part2}}
\end{figure*}

\begin{figure*}
\centering
\includegraphics[width=0.65\textwidth]{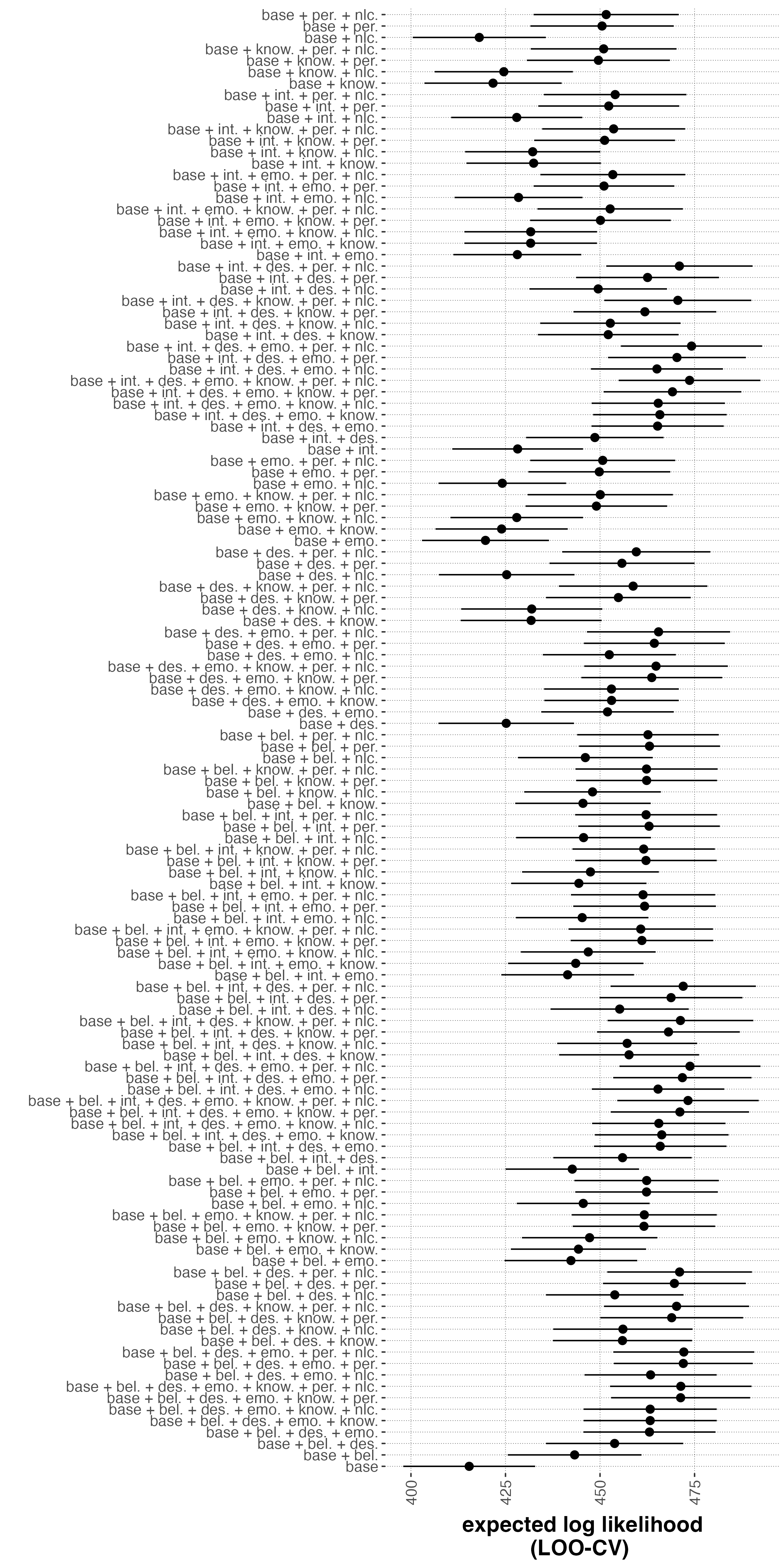}
\caption{Statistical comparison of models with different combinations of up to seven ATOMS predictors with LOO cross-validation, showing the estimated expected log-likelihoods for LOO-CV (x-axis) for different regression models (y-axis). Higher values indicate a model's better predictive fit to the behavioral accuracy on ToM datasets. All models contain baseline predictors model family, size, type and dataset type (summarized as ``base''). The ATOMS predictors are abbreviated (``bel.'' = belief, ``des.'' = desire, ``emo.'' = emotions, ``int.'' = intentions, ``know.'' = knowledge, ``per.'' = percepts, ``nlc.'' = non-literal communication).  The best model contains the predictors ``base + int. + des. + emo. + per. + nlc'', the worst model is the ``base'' only model.\label{app:fig:atoms-model-comparison}}
\end{figure*}


\section{Prompts for Stimulus Generation}
\label{app:sec:prompts-synthetic-loc-prompts}
For the generation of synthetic stimuli for the conditions FalseBelief, Desire, FalsePhotograph, HumanDescr, NonhumanDescr, MechInf, GameBelief, and MoralIntent, we used the following prompt:\newline

\noindent \textit{Please generate twenty more stories and questions following the structure of the given examples:\newline\newline----\newline\{examples\}}\newline

The examples consisted of five randomly selected items from the original or manually adjusted dataset, separated by "----" and each structured as follows:\newline

\noindent \textit{Story:\newline\{story\}\newline\newline Question:\newline\{question\}}\newline

For the conditions Deceptive, Ironic, Literal, and Meaningless, where the question is the same for all items, the following prompt was used:\newline

\noindent \textit{Please generate twenty more stories following the structure of the given examples:\newline\newline----\newline\{examples\}}\newline

The examples again were five randomly selected story items from the original dataset, separated by "----" and each prefixed by "Story:"

For GameOutcome, which is paired with GameBelief by using the same stories but different questions, we used the following prompt:\newline

\noindent \textit{For the given story, generate a question analogous to the examples.\newline\newline Examples:\newline\newline\{examples\}\newline\newline----\newline\newline Story:\newline\newline\{original story\}}\newline

Here, the examples were five randomly drawn items (each consisting of the story and one question) from the original GameOutcome data, and the original story was a single story item from the synthetic GameBelief data.

For DecisionOutcome, which is paired with moral intent by sharing the same overall setting (but not the whole narrative text), we used the following prompt:\newline

\noindent \textit{Convert the given story into a similar story (accompanied by a corresponding question) where no moral decision is involved, in analogy to the given examples.\newline\newline Examples:\newline\newline \{examples\}\newline\newline----\newline\newline Original story:\newline\newline \{original story\}}\newline

Here, the examples consisted of five triplets of a MoralIntent story and a DecisionOutcome story and question from the original or manually constructed data (introduced by "Original:", "New story:", and "Question:" respectively), and the original story was a single story item from the synthetic MoralIntent data.

\section{Localizer Suites}
\label{app:sec:localizer-suites}

\subsection{Details on each Localizer Suite}

\begin{enumerate}
    \item \textbf{LatentBeliefs}: to localize the ability to recognize latent mental action causes, we generate synthetic stimuli based on 6 items from \citet{SaxeKanwisher2003:People-thinking} where a person acts out of a false belief (henceforth called the FalseBelief condition) and on 6 items from the same paper where a person's desires do not match with the actual course of events (the Desire condition). The corresponding control stimuli are based on four other conditions from \citet{SaxeKanwisher2003:People-thinking} where no reasoning about mental causes is required: the FalsePhotograph condition involves photographs or other kinds of depictions that describe a state that does not exist anymore (i.e., false content), the HumanDescr condition involves physical descriptions of people (to contrast mental causes from mere mentioning of human attributes), the NonhumanDescr condition involves descriptions of non-human objects, and the MechInf conditions requires inferring a latent mechanical (i.e., non-mental) cause of an event. Since there are no questions given for the MechInf stories in \citet{SaxeKanwisher2003:People-thinking}, we manually add corresponding questions to align the format to that of the other conditions.\footnote{We only use a subset of 9 of the original mechanical inference stories where the actual hidden cause is unambiguous.}
    \item \textbf{CommunicativeIntent}: to localize the ability to recognize latent mental states in communication, we use synthetic stimuli based on data from \citet{bosco2017neural}. Following \citet{bosco2017neural}, we contrast a Deceptive condition, where an utterance is used in a deliberately misleading way, and an Ironic condition, where an utterance is used ironically, with control conditions where an utterance is meant literally (Literal) or where it is not related to the context at all (Meaningless).
    \item \textbf{GameBeliefs:} to localize the ability to reason about the latent causes for agent behavior in the specific setting of strategic games, we use synthetic stimuli based on data from \citet{chang2023mentalizing}. We generate synthetic stimuli for a GameBelief condition, where reasoning about the motivation for a specific decision in a trust game or ultimatum game is necessary, as well as paired GameOutcome stimuli, which concern the same game settings but merely involve the outcome of the game rather than the player's motivations. This suite is paired. 
    \item \textbf{MoralIntent}: to localize the ability to ascribe moral intent to actors, we generate synthetic stimuli based on 12 stimuli from \citet{young2007neural}, where a person chooses to perform an action despite being aware of its harmful consequences for another actor (MoralIntent). We base our synthetic control stimuli on manually created paired control items, where a decision accidentally leads to a harmful outcome for the acting person themselves (DecisionOutcome). This suite is paired.
\end{enumerate}

Table~\ref{tab:LocalizerSuitesAll} includes one example stimulus for each condition, and Table~\ref{tab:app:localizer-atoms} contains the ATOMS categories (on a coarse-grained level) addressed by the localization target conditions.

\begin{table*}[t]
    \footnotesize
    \centering
    \begin{tabularx}{\linewidth}{l X}
        \toprule
         %
         \multicolumn{2}{l}{\textbf{LatentBeliefs} \citep{SaxeKanwisher2003:People-thinking}} \\
         FalseBelief (trgt)  & Story: Tom dropped his keys in the bowl by the door before going for a run. While he was out, Mia put the keys into the desk drawer to tidy up. Tom came back a bit later. Statement / question: Tom will first look for his keys in the ($\bullet \text{ bowl} \bullet \text{ drawer}$) \\
         Desire (trgt)       & Story: Coach Ramirez booked the soccer field for Saturday practice. He told the team they’d be outside unless lightning forced them into the gym. A storm never came, and they kept the plan. Statement / question: The coach wanted to practice ($\bullet \text{ indoors} \bullet \text{ outdoors}$) \\
         FalsePhoto (ctrl)   & Story: A bus schedule was printed last month. Yesterday the transit agency changed Route 9 to depart an hour later. All the old paper schedules are still circulating. Statement / question: On the printed schedule, Route 9 departs ($\bullet \text{ earlier} \bullet \text{ later}$) \\
         HumanDescr (ctrl)   & Story: Nora never goes out in the rain without her favorite outfit. She wears a bright yellow raincoat and matte black rubber boots. Even her umbrella is speckled with tiny white dots. Statement / question: Nora's raincoat is ($\bullet \text{ yellow} \bullet \text{ black}$) \\
         MechInf (ctrl)      & Story: She forgot a full glass bottle of soda in the freezer overnight. By morning, the freezer door was stuck and there were shards of glass around the bottle. The cap was still on, but the liquid had turned to ice and expanded. Statement / question: The bottle shattered because ($\bullet \text{ freezing water expanded} \bullet \text{ a thief broke in}$) \\
         NonhumDescr (ctrl) & Story: Penguins are birds that cannot fly but swim swiftly. They use their stiff wings like flippers to push through the water. A layer of fat keeps them warm in icy seas. Statement / question: In the water, penguins propel themselves mainly with their ($\bullet \text{ wings} \bullet \text{ tails}$) \\
         \midrule
         \multicolumn{2}{l}{\textbf{CommunicativeIntent} \citep{bosco2017neural}} \\
         Deceptive (trgt)    & Story: Oliver has been invited to his manager’s poetry reading. He finds the poems boring and overwrought, but he wants to make a good impression at work. After the reading, his manager asks what he thought. Oliver replies: “It was deeply moving, I loved every minute” Statement / question: Was this ($\bullet \text{ literal} \bullet \text{ deceitful} \bullet \text{ ironic} \bullet \text{ meaningless?}$) \\
         Ironic (trgt)       & Story: Luca and Giulia are driving to a wedding. They hit a standstill on the ring road, horns blaring and heat shimmering off the tarmac. The GPS suddenly adds another hour to their arrival time even though the venue is close by. Giulia looks at the bouquet wilting on the dashboard. Luca sighs and says: “Great, we’ll be early” Statement / question: Was this ($\bullet \text{ literal} \bullet \text{ deceitful} \bullet \text{ ironic} \bullet \text{ meaningless?}$) \\

         Literal (ctrl)      & Story: Elena is making toast for breakfast but accidentally burns it black. Her partner Diego walks in and raises an eyebrow. Elena answers: "I'll make another one." Statement / question: Was this ($\bullet \text{ literal} \bullet \text{ deceitful} \bullet \text{ ironic} \bullet \text{ meaningless?}$) \\

         Meaningless (ctrl)  & Story: Liam and Nora are doing the weekly shop at the supermarket. They’ve already picked up milk, bread, and fruit. Liam reaches for a carton and asks Nora if they still have eggs at home. Nora answers: “The station is two blocks away” Statement / question: Was this ($\bullet \text{ literal} \bullet \text{ deceitful} \bullet \text{ ironic} \bullet \text{ meaningless?}$) \\
         \midrule
         \multicolumn{2}{l}{\textbf{GameBeliefs\textsuperscript{*}} \citep{chang2023mentalizing}} \\
         GameBelief (trgt)   & Story: Nora and Ben play an Ultimatum Game. Nora is the proposer and receives 20 Euros to divide between her and another player. She intends to offer half to Ben, but a glitch records her offer as 1 Euro. Ben sees 1 Euro on his screen, frowns, and rejects immediately. Statement / question: Ben evidently believes that ($\bullet \text{ Nora is trying to cheat him} \bullet \text{ a glitch reduced the offer}$) \\
         GameOutcome (ctrl)  & Story: Nora and Ben play an Ultimatum Game. Nora is the proposer and receives 20 Euros to divide between her and another player. She intends to offer half to Ben, but a glitch records her offer as 1 Euro. Ben sees 1 Euro on his screen, frowns, and rejects immediately. Statement / question: Ben will now have ($\bullet \text{ 1 Euro} \bullet \text{ 0 Euro}$) \\
         \midrule
         \multicolumn{2}{l}{\textbf{MoralIntent\textsuperscript{*}} \citep{young2007neural}} \\
         MoralIntent (trgt)  & Story: Anna and her friend are hiking along a cliffside trail when they reach a narrow rock ledge overlooking a canyon. The ledge is unstable and will crumble if someone steps onto it, causing a deadly fall. Because Anna sees deep cracks running through the rock and read a posted warning about recent collapses, she believes the ledge is unsafe and stepping on it could be fatal. Anna urges her friend to step onto the ledge for a better view; the ledge collapses and her friend falls to her death. Statement / question: Urging her friend onto the ledge was ($\bullet \text{ permissible} \bullet \text{ forbidden}$) \\
         DecOutcome (ctrl)   & Story: Anna and her friend are hiking along a cliffside trail when they reach a narrow rock ledge overlooking a canyon. The ledge is unstable and will crumble if someone steps onto it, causing a deadly fall. While trying to take a photo of the view, Anna steps onto the ledge; it collapses, and she falls to her death. Statement / question: Stepping onto the ledge was ($\bullet \text{ a good idea} \bullet \text{ a bad idea}$) \\
        \bottomrule
    \end{tabularx}
    \caption{Examples of synthetic GPT-5 generated stimuli for all localizer suites. Dots indicate answer options. The references indicate the source of the original stimuli that were used to prompt GPT-5. 
    }
    \label{tab:LocalizerSuitesAll}
\end{table*}

\begin{table*}
\footnotesize
    \begin{tabular}{c|c|c|c|c|c|c|c}
         Dataset & beliefs & intentions & desires & emotions & knowledge & percepts & NLC \\
         LatentBelief \citep{SaxeKanwisher2003:People-thinking} & & & & & & & \\
        $ \bullet\text{ FalseBelief} $& \checkmark & $\times$ & $\times$ & $\times$ & $\times$ & \checkmark & $\times$ \\
         $\bullet\text{ Desire}$ & $\times$ & \checkmark & \checkmark & $\times$ & $\times$ & $\times$ & $\times$ \\
         CommunicativeIntent \citep{bosco2017neural} & & & & & & & \\
         $\bullet\text{ Deceptive}$ & $\times$ & \checkmark & $\times$ & $\times$ & $\times$ & $\times$ & \checkmark \\
         $\bullet\text{ Irony}$ & $\times$ & \checkmark & $\times$ & $\times$ & $\times$ & $\times$ & \checkmark \\
         MoralIntent \citep{chang2023mentalizing} & $\times$ & \checkmark & $\times$ & $\times$ & \checkmark & $\times$ & $\times$ \\
         GameBelief \citep{young2007neural} & \checkmark & $\times$ & \checkmark & \checkmark & $\times$ & \checkmark & $\times$ \\
    \end{tabular}
    \caption{Classification of localizer suites according to the ATOMS framework \citep{beaudoin2020systematic}. NLC stands for ``non-literal communication''. \label{tab:app:localizer-atoms}}
\end{table*}

\subsection{Formatting in Localization Procedures}
\label{sec:app:localizer-instructions-formatting}
For all localizer suites, the following instruction was prepended to each item before they were passed through the model and the activations were retrieved: \\

\noindent \textit{In this experiment, you will read a story, a question or a statement and select the best answer among the provided options.} \\

Additionally, for the GameBeliefs localizer, the following instructions were appended to the general instruction (selected to match the game mentioned in each particular item): \\

\noindent Ultimatum game: \textit{The story is about the Ultimatum Game. There are two players who divide a sum of money in this game. The first player, the proposer, proposes a division of the sum with the second player, the responder. The responder can either accept the proposed division or reject it. If the responder accepts, the money is split according to the proposal; if the responder rejects, neither player receives anything.} \newline \\
\noindent Trust game: \textit{The story is about the Trust Game. There are two players who play an investment game. Both players are given some quantity of money. The first player is told that they must send some amount of their money to an anonymous second player, though the amount sent may be zero. The first player is also informed that whatever they send will be tripled by the experimenter and given to the second player. The second player is then told to also give some amount of the now-tripled money back to the first player, even if that amount is zero.} \\

The localization stimuli were formatted into the following template, resulting in the final language model input: \textit{\{instruction\}Story: \{story\}\textbackslash n Statement / question: \{question\} \textbackslash n Options:\textbackslash n\{options\}\textbackslash n}, where the answer options were formatted as a bullet point list with dashes. 

For all language models that expect a special template for formatting inputs, i.e., when the corresponding tokenizer provides a chat template, the inputs were put into the chat template as the user turn of the conversation. 
Then, an assistant turn was appended with the following content: \textit{``Answer: \{answer prefix\}''}. 
The answer prefix contained the sentence that was to be completed with the answer options of the questions, but without any answers.
The parameter to continue the final message was set to True.
The last token at which the activations were retrieved was therefore set up to, ideally, concentrate representations of the target latent information required to answer the question.

\subsection{Principal Component Analyses of Synthetic Localizers}
\label{sec:app:localizer-pca}
\begin{figure*}
\includegraphics[width=\textwidth]{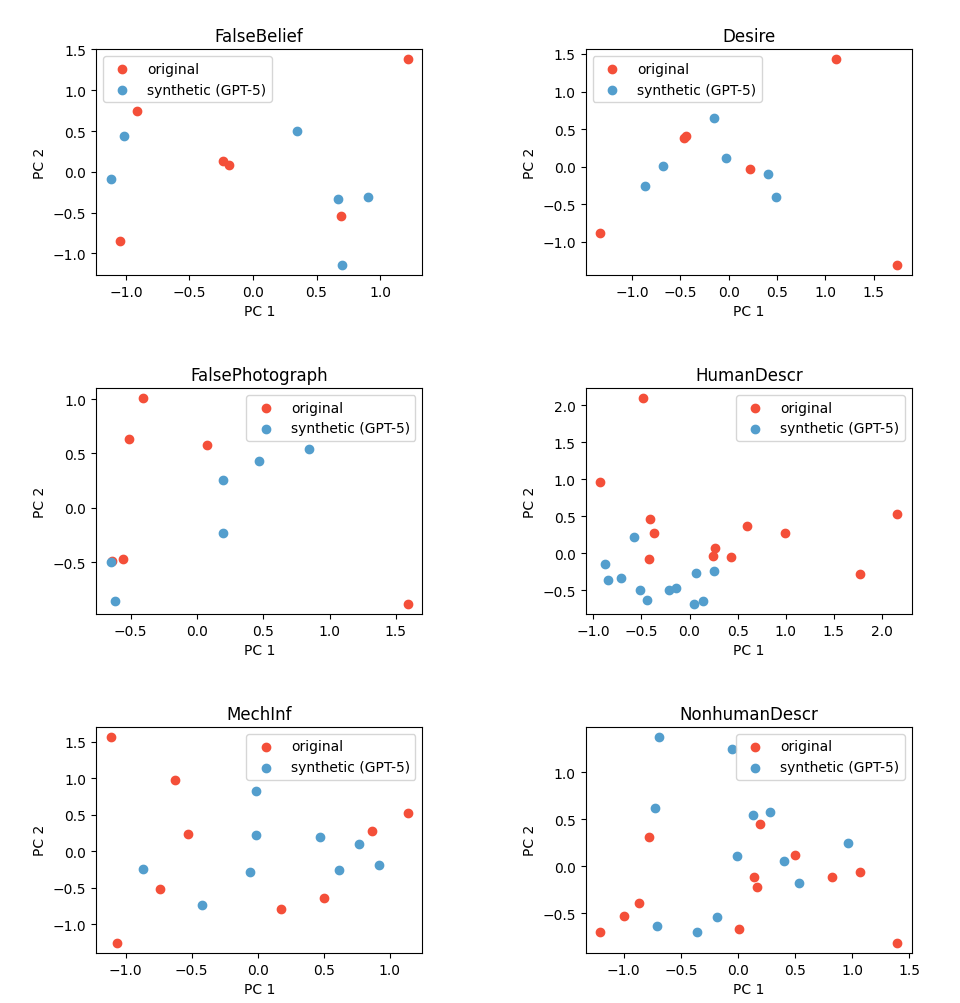}
\caption{PCA with two components of sentence embeddings (from SentenceTransformer's all-MiniLM-L6-v2 model) of the original stimuli and an equal number of randomly selected synthetic stimuli, for the conditions FalseBelief, Desire, FalsePhotograph, HumanDescr, MechInf, and NonhumanDescr (all from the LatentBeliefs suite).
\label{app:fig:pca1}}
\end{figure*}

\begin{figure*}
\includegraphics[width=\textwidth]{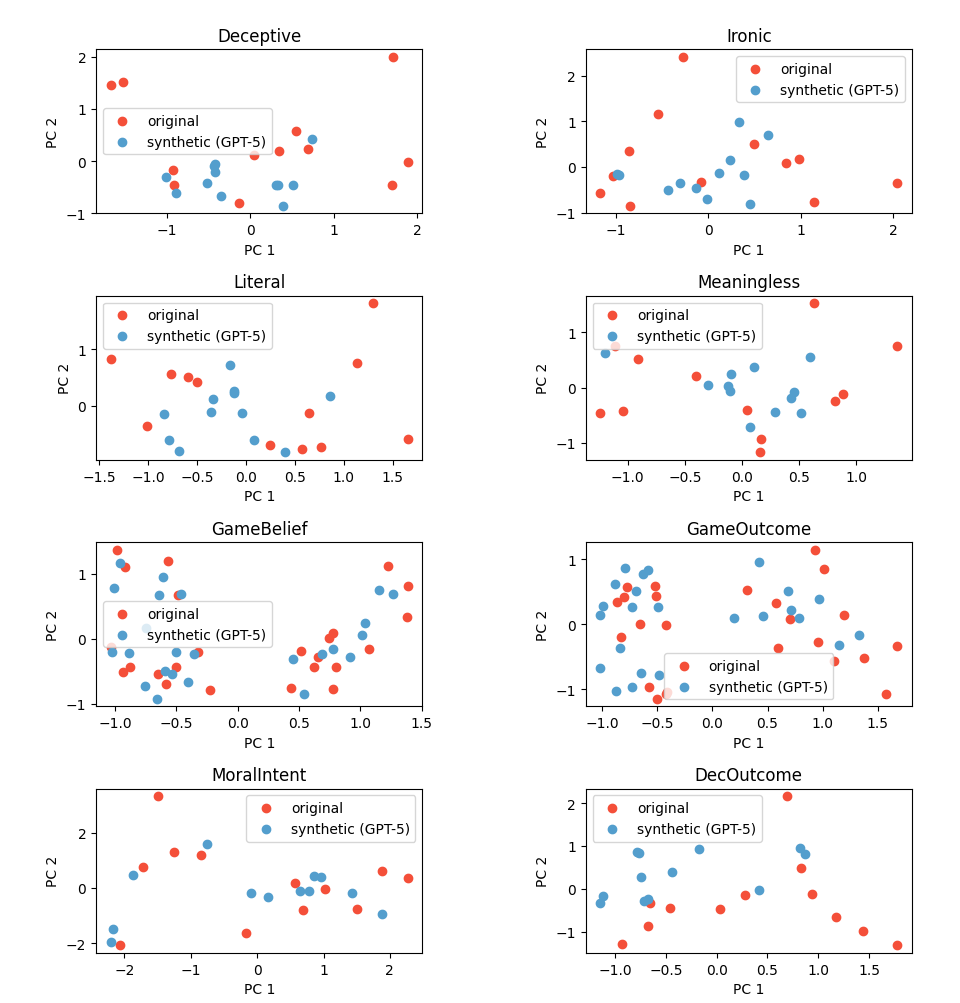}
\caption{PCA with two components of sentence embeddings (from SentenceTransformer's all-MiniLM-L6-v2 model) of the original stimuli and an equal number of randomly selected synthetic stimuli, for the conditions Deceptive, Ironic, Literal, Meaningless (from the CommunicativeIntent suite), GameBelief, GameOutcome (from GameBeliefs), MoralIntent, DecOutcome (from MoralIntent, respectively).
\label{app:fig:pca2}}
\end{figure*}

To assess the GPT-5 generated localizer stimuli in comparison to original localizer materials from the literature in terms of surface-level and statistical properties to which LMs might be particularly sensitive, a Principal Component Analysis (PCA) with two components was performed.
For each condition of each suite, sentence embeddings for the story part (i.e., the context narrative without the question or answer options) of all original stimuli and an equal number of randomly selected synthetic stimuli were calculated using the SentenceTransformer's all-MiniLM-L6-v2 model \citep{reimers-2019-sentence-bert}. 
The results of the PCA analysis of the original and synthetic data for each condition are showing in Figures~\ref{app:fig:pca1}--\ref{app:fig:pca2}. Visual inspection confirms that original and synthetic stimuli do not cluster separately, indicating that there is no systematic mismatch between them. 

Only for one of the conditions (HumanDescr from the LatentBeliefs suite), the two groups can clearly be separated; manual inspection suggests that the difference lies in the fact that the original stimuli describe the physical appearance of a person only, while the synthetic stimuli include other characteristics such as habits. Since the task at hand (i.e., answering simple questions about the described person) does not require Theory-of-Mind reasoning in either case, this does not affect using the synthetic stimuli as a non-ToM control condition. For two other conditions (FalsePhotograph from the LatentBeliefs suite and DecOutcome from the MoralIntent suite), some weak clustering is visible, but original and synthetic stimuli are not clearly separated, and the amount of clustering differs strongly between different samples of synthetic stimuli. Finally, for GameBelief and GameOutcome from the GameBeliefs suite, the sentence embeddings form two clusters, both including a similar number of original and synthetic stimuli; this can be explained by the presence of two types of strategic games in the dataset (trust games and ultimatum games).

\section{Subnetwork Localization Procedure Details}



\label{app:sec:loc-models-datasets}
We select the following 20 models for our localization experiments.

\noindent \textbf{Llama}: Llama-3.1-8B, Llama-2-13b-hf, Llama-2-13b-chat-hf, Llama-3.1-8B-Instruct, Llama-3.1-70B, Llama-3.3-70B-Instruct

\noindent \textbf{Falcon}: Falcon3-7B, Falcon3-7B-Instruct, Falcon3-10B, Falcon3-10B-Instruct

\noindent \textbf{Gemma}: gemma-2-9b, gemma-2-9b-it, gemma-2-27b, gemma-2-27b-it

\noindent \textbf{Qwen}: Qwen2.5-7B, Qwen2.5-7B-Instruct, Qwen2.5-32B, Qwen2.5-32B-Instruct, Qwen2.5-72B, Qwen2.5-72B-Instruct

In the ablation experiments (Section~\ref{sec:ablation-experiments}), each model was evaluated on BLiMP \citep{warstadt-etal-2020-blimp-benchmark} and SNLI \citep{bowman2015large}, and on the following selected ToM and pragmatics datasets on which the intact models showed an above-chance performance (described in detail above in Section~\ref{sec:app:datasets}):

\begin{itemize}
    \item Llama-3.1-8B: BigToM (forward),
            EmoBench (emotion application),
            EmoBench (understanding emotion),
            EPITOME (recursive mindreading),
            EWoK (agent properties),
            EWoK (social interactions),
            FauxPasEAI (Q1 \& Q4),
            HuFloyd (deceits),
            HuFloyd (indirect speech),
            HuFloyd (irony),
            HuFloyd (Gricean maxims),
            HuFloyd (metaphor),
            IMPPRES (implicature),
            IMPPRES (presupposition),
            LUDWIG,
            OpenToM (attitude),
            PUB (agreement detection),
            PUB (deictic QA),
            PUB (indirectness classification),
            PUB (understanding sarcasm),
            SocialIQA,
            Triangle-COPA
    \item Llama-3.1-8B-Instruct: BigToM (forward),
            EmoBench (emotion application),
            EmoBench (understanding emotion),
            EPITOME (recursive mindreading),
            EWoK (agent properties),
            EWoK (social interactions),
            FauxPasEAI (Q1 \& Q4),
            HuFloyd (deceits),
            HuFloyd (indirect speech),
            HuFloyd (irony),
            HuFloyd (Gricean maxims),
            HuFloyd (metaphor),
            IMPPRES (implicature),
            IMPPRES (presupposition),
            OpenToM (attitude),
            PUB (agreement detection),
            PUB (deictic QA),
            PUB (understanding sarcasm),
            SocialIQA,
            ToMi (first-order false-belief),
            Triangle-COPA
    \item Llama-2-13b-hf: BigToM (forward),
            EmoBench (emotion application),
            EPITOME (recursive mindreading),
            EWoK (agent properties),
            EWoK (social interactions),
            HuFloyd (deceits),
            HuFloyd (indirect speech),
            HuFloyd (irony),
            HuFloyd (Gricean maxims),
            HuFloyd (metaphor),
            IMPPRES (presupposition),
            LUDWIG,
            OpenToM (attitude),
            PUB (agreement detection),
            PUB (deictic QA),
            PUB (understanding sarcasm),
            SocialIQA,
            ToMi (first-order false-belief),
            Triangle-COPA
    \item Llama-2-13b-chat-hf: BigToM (forward),
            EmoBench (emotion application),
            EmoBench (understanding emotion),
            EPITOME (recursive mindreading),
            EWoK (agent properties),
            EWoK (social interactions),
            HuFloyd (deceits),
            HuFloyd (indirect speech),
            HuFloyd (irony),
            IMPPRES (implicature),
            IMPPRES (presupposition),
            LUDWIG,
            OpenToM (attitude),
            PUB (agreement detection),
            PUB (deictic QA),
            PUB (indirectness classification),
            PUB (understanding sarcasm),
            SocialIQA,
            Triangle-COPA 
    \item Llama-3.1-70B: BigToM (forward),
            EmoBench (emotion application),
            EmoBench (understanding emotion),
            EPITOME (recursive mindreading),
            EWoK (agent properties),
            EWoK (social interactions),
            HuFloyd (deceits),
            HuFloyd (indirect speech),
            HuFloyd (irony),
            HuFloyd (Gricean maxims),
            HuFloyd (metaphor),
            IMPPRES (implicature),
            LUDWIG,
            PUB (agreement detection),
            PUB (deictic QA),
            PUB (indirectness classification),
            PUB (understanding sarcasm),
            SocialIQA,
            Triangle-COPA
    \item Llama-3.3-70B-Instruct: BigToM (forward),
            EmoBench (emotion application),
            EmoBench (understanding emotion),
            EPITOME (recursive mindreading),
            EWoK (agent properties),
            EWoK (social interactions),
            FauxPasEAI (Q1 \& Q4),
            HuFloyd (deceits),
            HuFloyd (indirect speech),
            HuFloyd (irony),
            HuFloyd (Gricean maxims),
            HuFloyd (metaphor),
            IMPPRES (implicature),
            IMPPRES (presupposition),
            LUDWIG,
            OpenToM (attitude),
            PUB (agreement detection),
            PUB (deictic QA),
            PUB (indirectness classification),
            PUB (understanding sarcasm),
            SocialIQA,
            Triangle-COPA
    \item Falcon3-7B: BigToM (forward),
            EmoBench (emotion application),
            EmoBench (understanding emotion),
            EPITOME (recursive mindreading),
            EWoK (agent properties),
            EWoK (social interactions),
            FauxPasEAI (Q1 \& Q4),
            HuFloyd (deceits),
            HuFloyd (indirect speech),
            HuFloyd (irony),
            HuFloyd (Gricean maxims),
            HuFloyd (metaphor),
            IMPPRES (implicature),
            IMPPRES (presupposition),
            LUDWIG,
            OpenToM (attitude),
            PUB (agreement detection),
            PUB (deictic QA),
            PUB (indirectness classification),
            PUB (understanding sarcasm),
            SocialIQA,
            Triangle-COPA
    \item Falcon3-7B-Instruct: BigToM (forward),
            EmoBench (emotion application),
            EmoBench (understanding emotion),
            EPITOME (recursive mindreading),
            EWoK (agent properties),
            EWoK (social interactions),
            FauxPasEAI (Q1 \& Q4),
            HuFloyd (deceits),
            HuFloyd (indirect speech),
            HuFloyd (irony),
            HuFloyd (Gricean maxims),
            HuFloyd (metaphor),
            IMPPRES (implicature),
            IMPPRES (presupposition),
            OpenToM (attitude),
            PUB (indirectness classification),
            PUB (understanding sarcasm),
            SocialIQA,
            Triangle-COPA
    \item Falcon3-10B: BigToM (forward),
            EmoBench (emotion application),
            EmoBench (understanding emotion),
            EPITOME (recursive mindreading),
            EWoK (agent properties),
            EWoK (social interactions),
            FauxPasEAI (Q1 \& Q4),
            HuFloyd (deceits),
            HuFloyd (indirect speech),
            HuFloyd (irony),
            HuFloyd (Gricean maxims),
            HuFloyd (metaphor),
            IMPPRES (implicature),
            IMPPRES (presupposition),
            LUDWIG,
            OpenToM (attitude),
            PUB (agreement detection),
            PUB (deictic QA),
            PUB (indirectness classification),
            PUB (understanding sarcasm),
            SocialIQA,
            Triangle-COPA
    \item Falcon3-10B-Instruct: BigToM (forward),
            EmoBench (emotion application),
            EmoBench (understanding emotion),
            EPITOME (recursive mindreading),
            EWoK (agent properties),
            EWoK (social interactions),
            FauxPasEAI (Q1 \& Q4),
            HuFloyd (deceits),
            HuFloyd (indirect speech),
            HuFloyd (irony),
            HuFloyd (Gricean maxims),
            HuFloyd (metaphor),
            IMPPRES (implicature),
            IMPPRES (presupposition),
            OpenToM (attitude),
            PUB (deictic QA),
            PUB (indirectness classification),
            PUB (understanding sarcasm),
            SocialIQA,
            Triangle-COPA
    \item all gemma models: ToMi (first-order false-belief), 
            PUB (agreement detection),
            PUB (indirectness classification), 
            FauxPasEAI (Q1 \& Q4),
            LUDWIG,
            OpenToM (attitude), 
            HuFloyd (irony),
            IMPPRES (presupposition),
            PUB (understanding sarcasm),
            PUB (deictic QA), 
            HuFloyd (Gricean maxims),
            HuFloyd (metaphor),
            IMPPRES (implicature),
            EmoBench (understanding emotion),
            Triangle-COPA,
            EWoK (agent properties),
            HuFloyd (indirect speech),
            HuFloyd (deceits),
            BigToM (forward),
            EWoK (social interactions),
            SocialIQA,
            EPITOME (recursive mindreading),
            EmoBench (emotion application)
    \item Qwen2.5-7B: BigToM (forward),
            EmoBench (emotion application),
            EmoBench (understanding emotion),
            EPITOME (recursive mindreading),
            EWoK (agent properties),
            EWoK (social interactions),
            FauxPasEAI (Q1 \& Q4),
            HuFloyd (deceits),
            HuFloyd (indirect speech),
            HuFloyd (Gricean maxims),
            HuFloyd (metaphor),
            IMPPRES (implicature),
            IMPPRES (presupposition),
            LUDWIG,
            OpenToM (attitude),
            PUB (agreement detection),
            PUB (indirectness classification),
            PUB (understanding sarcasm),
            SocialIQA,
            Triangle-COPA
    \item Qwen2.5-7B-Instruct: BigToM (forward),
            EmoBench (emotion application),
            EmoBench (understanding emotion),
            EPITOME (recursive mindreading),
            EWoK (agent properties),
            EWoK (social interactions),
            FauxPasEAI (Q1 \& Q4),
            HuFloyd (deceits),
            HuFloyd (indirect speech),
            HuFloyd (Gricean maxims),
            HuFloyd (metaphor),
            IMPPRES (implicature),
            IMPPRES (presupposition),
            LUDWIG,
            PUB (deictic QA),
            PUB (understanding sarcasm),
            SocialIQA,
            Triangle-COPA
    \item Qwen2.5-32B: BigToM (forward),
            EmoBench (emotion application),
            EmoBench (understanding emotion),
            EPITOME (recursive mindreading),
            EWoK (agent properties),
            EWoK (social interactions),
            FauxPasEAI (Q1 \& Q4),
            HuFloyd (deceits),
            HuFloyd (indirect speech),
            HuFloyd (irony),
            HuFloyd (Gricean maxims),
            HuFloyd (metaphor),
            IMPPRES (implicature),
            IMPPRES (presupposition),
            LUDWIG,
            OpenToM (attitude),
            PUB (agreement detection),
            PUB (deictic QA),
            PUB (indirectness classification),
            PUB (understanding sarcasm),
            SocialIQA,
            Triangle-COPA
    \item Qwen2.5-32B-Instruct: BigToM (forward),
            EmoBench (emotion application),
            EmoBench (understanding emotion),
            EPITOME (recursive mindreading),
            EWoK (agent properties),
            EWoK (social interactions),
            FauxPasEAI (Q1 \& Q4),
            HuFloyd (deceits),
            HuFloyd (indirect speech),
            HuFloyd (irony),
            HuFloyd (Gricean maxims),
            HuFloyd (metaphor),
            IMPPRES (implicature),
            IMPPRES (presupposition),
            LUDWIG,
            PUB (deictic QA),
            SocialIQA,
            Triangle-COPA
    \item Qwen2.5-72B: BigToM (forward),
            EmoBench (emotion application),
            EmoBench (understanding emotion),
            EPITOME (recursive mindreading),
            EWoK (agent properties),
            EWoK (social interactions),
            FauxPasEAI (Q1 \& Q4),
            HuFloyd (deceits),
            HuFloyd (indirect speech),
            HuFloyd (irony),
            HuFloyd (Gricean maxims),
            HuFloyd (metaphor),
            IMPPRES (implicature),
            IMPPRES (presupposition),
            LUDWIG,
            OpenToM (attitude),
            PUB (agreement detection),
            PUB (deictic QA),
            PUB (indirectness classification),
            PUB (understanding sarcasm),
            SocialIQA,
            Triangle-COPA
    \item Qwen2.5-72B-Instruct: BigToM (forward),
            EmoBench (emotion application),
            EmoBench (understanding emotion),
            EPITOME (recursive mindreading),
            EWoK (agent properties),
            EWoK (social interactions),
            FauxPasEAI (Q1 \& Q4),
            HuFloyd (deceits),
            HuFloyd (indirect speech),
            HuFloyd (irony),
            HuFloyd (Gricean maxims),
            HuFloyd (metaphor),
            IMPPRES (implicature),
            IMPPRES (presupposition),
            LUDWIG,
            OpenToM (attitude),
            PUB (agreement detection),
            PUB (deictic QA),
            PUB (indirectness classification),
            PUB (understanding sarcasm),
            SocialIQA,
            Triangle-COPA
\end{itemize}


\begin{figure*}
\includegraphics[width=\textwidth]{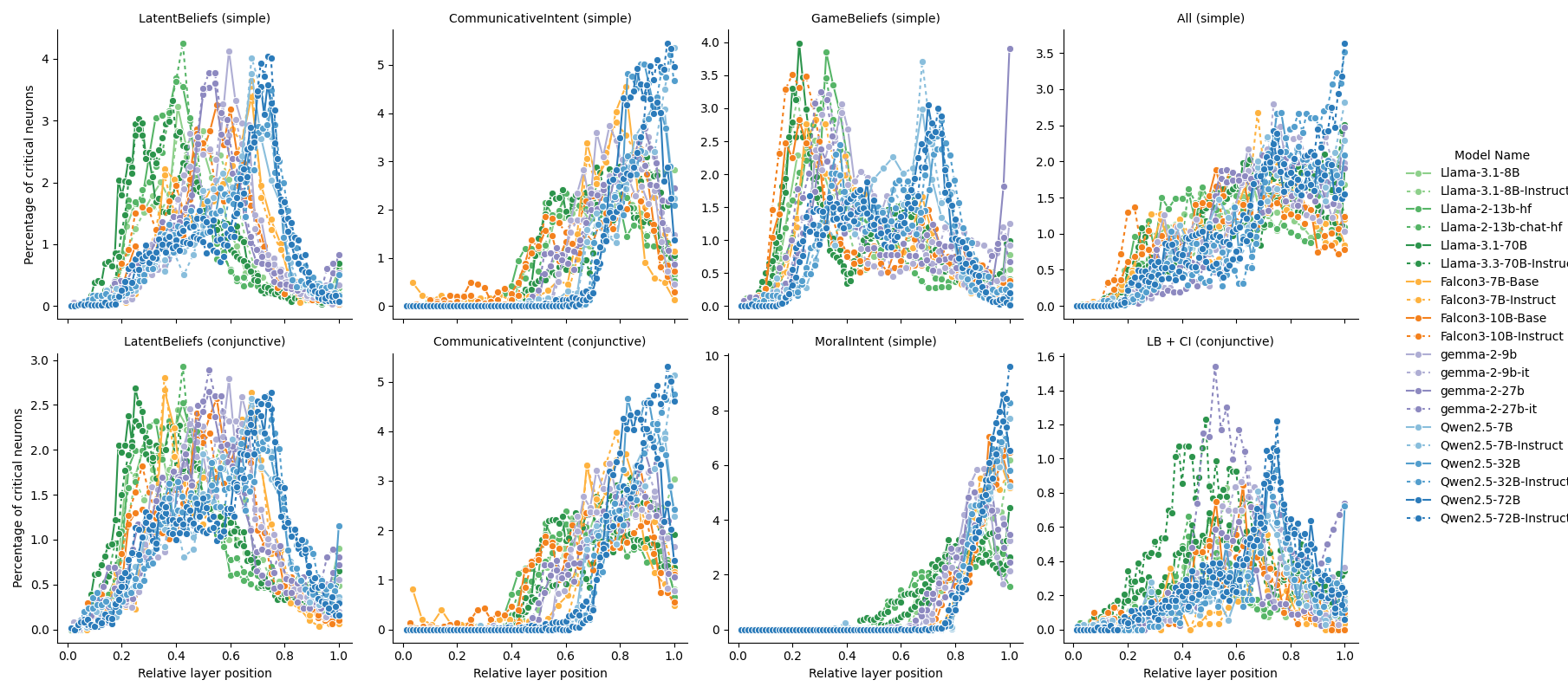}
\caption{Distribution of top 1\% of most active units across the model layers, identified with respect to the synthetically generated  ToM localizer suite through last-token pooling. The mask `theory-of-mind' identified most significant units under simple selection, the mask `theory-of-mind-conjunctive' identified most significant units under conjunctive selection, and the random masks identified the least significant units with respect to the same selection approaches, respectively. 
\label{app:fig:loc-tom-synthetic-critical}}
\end{figure*}
\begin{figure*}
\includegraphics[width=\textwidth]{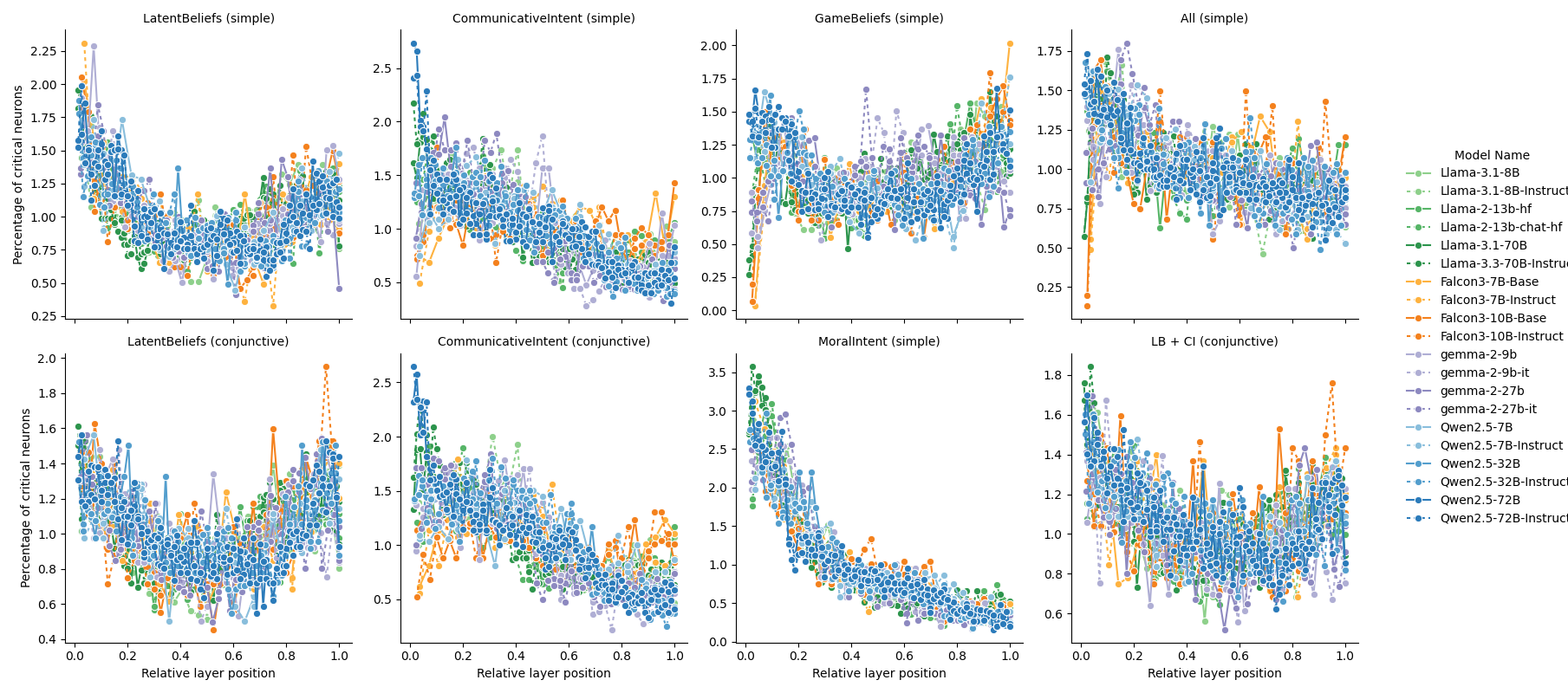}
\caption{Distribution of top 1\% of least active units across the model layers, identified with respect to the synthetically generated  ToM localizer suite through last-token pooling. The mask `theory-of-mind' identified most significant units under simple selection, the mask `theory-of-mind-conjunctive' identified most significant units under conjunctive selection, and the random masks identified the least significant units with respect to the same selection approaches, respectively. 
\label{app:fig:loc-tom-synthetic-random}}
\end{figure*}


\begin{figure*}[h]
\centering
\includegraphics[width=0.98\textwidth]{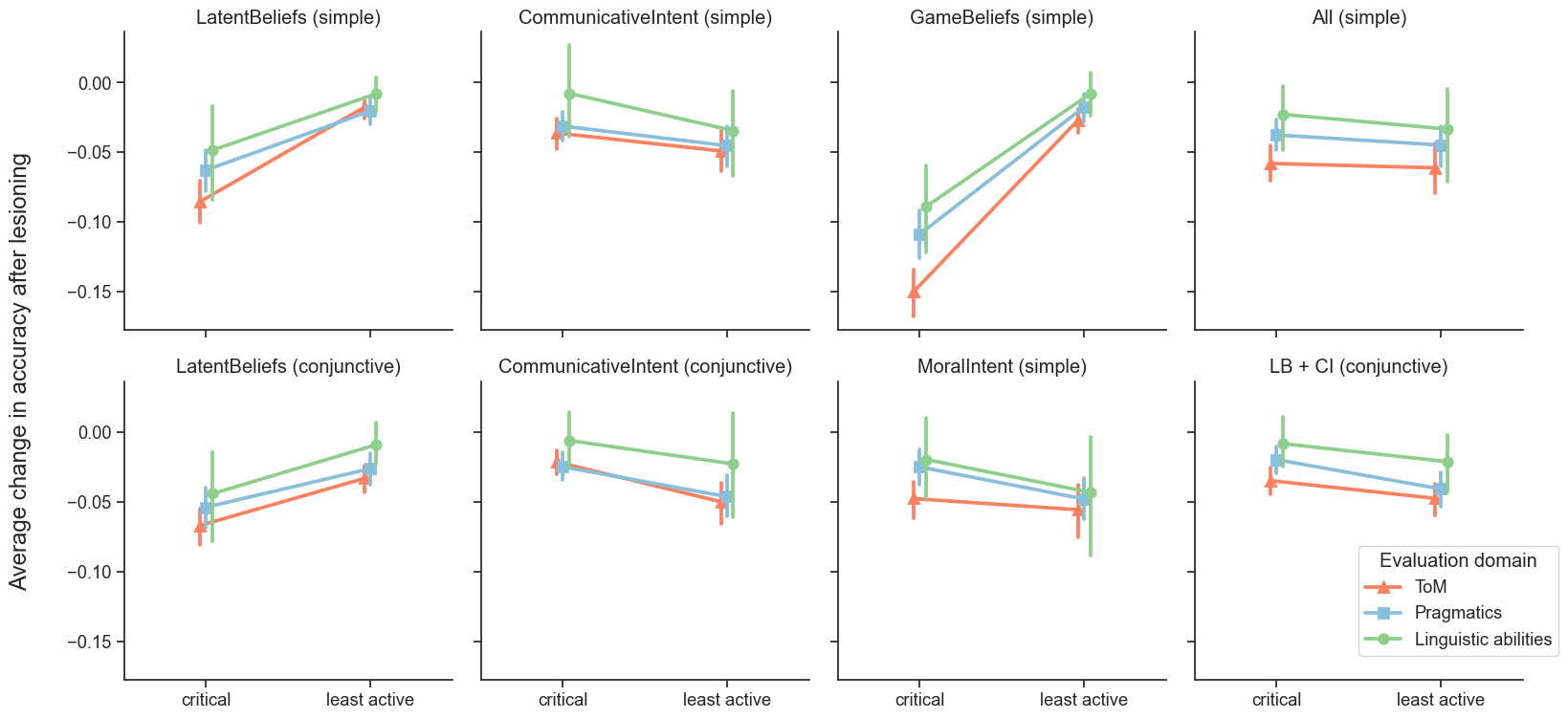}
\caption{Average difference in accuracy, by localizer, across models. \label{fig:loc-ablation-acc-acrossModels}}
\end{figure*}
\begin{figure*}[h]
\includegraphics[width=0.95\textwidth]{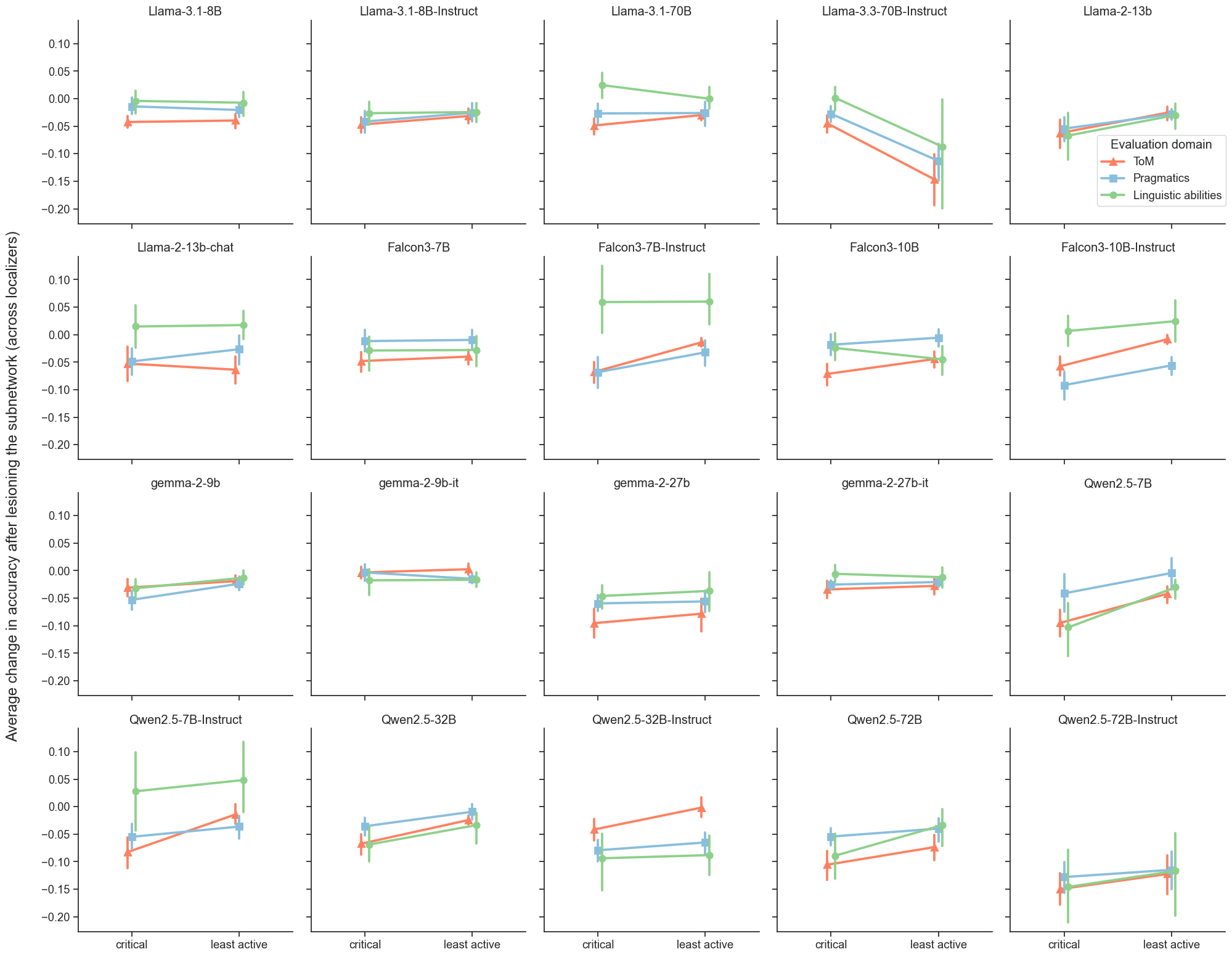}
\caption{Average change in accuracy (y-axis) relative to the intact model when applying ablations of the target or least active networks (x-axis) across evaluation datasets and localizer suites for different evaluated models (facets). A stronger decrease in accuracy for the critical ablation on target domains (ToM and pragmatics, color) is better. 95\% CIs indicate change in by-dataset accuracy across localizer suites. \label{fig:loc-ablation-acc}
}
\end{figure*}
\begin{figure*}
\includegraphics[width=\textwidth]{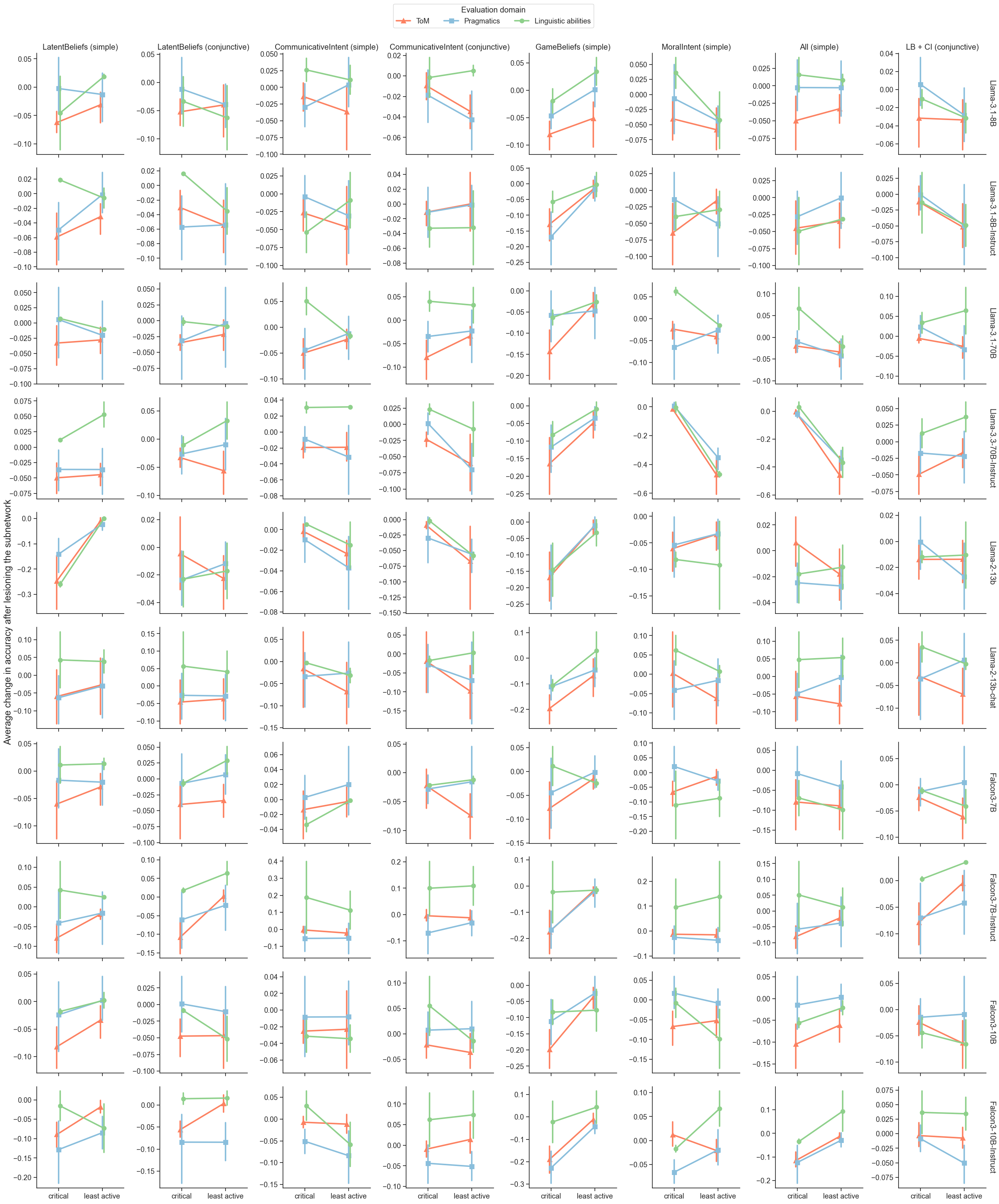}
\caption{Average change in accuracy across evaluation datasets (y-axis) when different localizer datasets are used (columns) in Llama and Falcon models (rows) when ablating the critical~vs.~the least active subnetworks (x-axis).  
\label{app:fig:loc-ablation-acc-byLocalizer-part1}}
\end{figure*}

\begin{figure*}
\includegraphics[width=\textwidth]{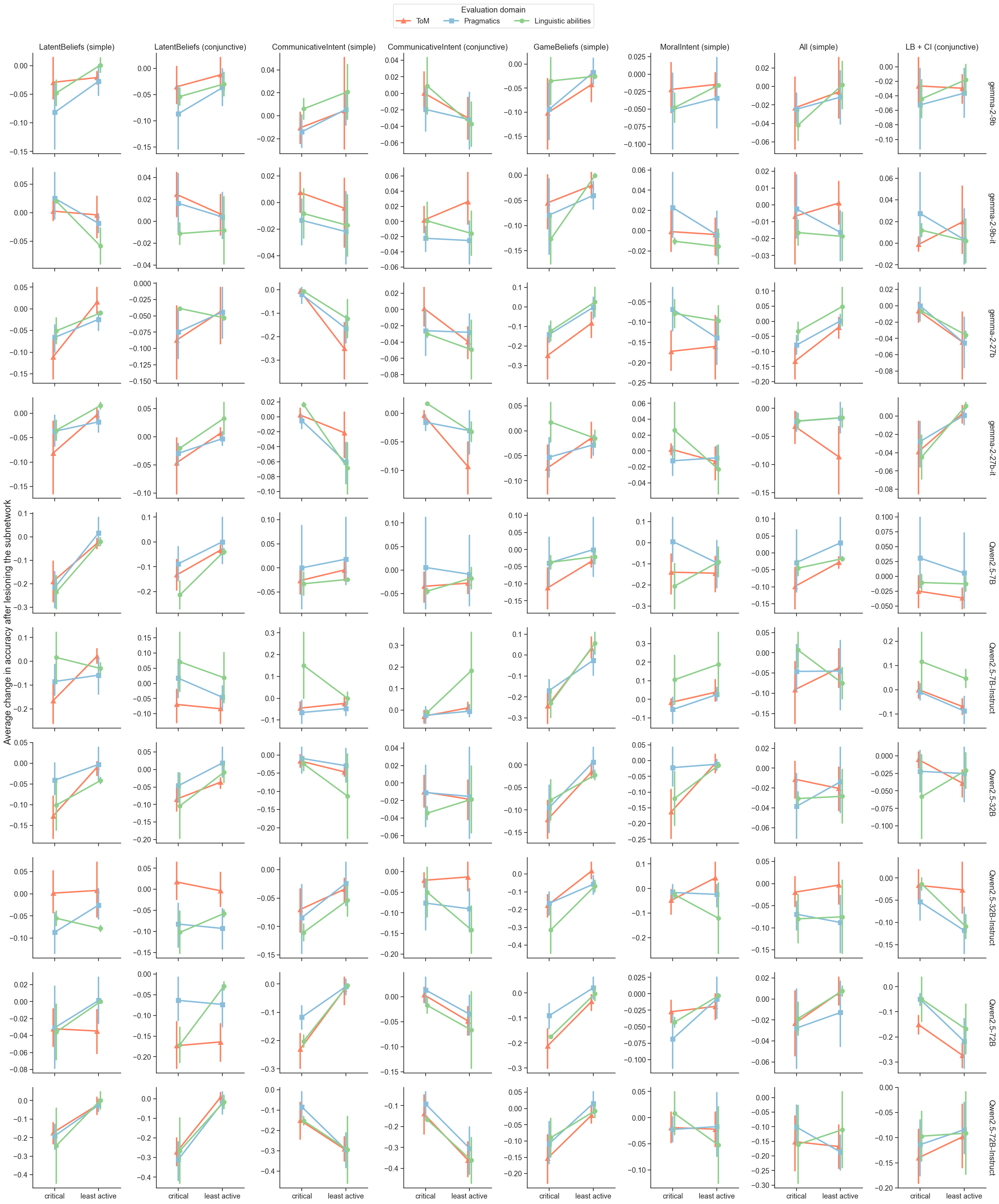}
\caption{Average change in accuracy across evaluation datasets (y-axis) when different localizer datasets are used (columns) in Gemma and Qwen models (rows) when ablating the critical~vs.~the least active subnetworks (x-axis). 
\label{app:fig:loc-ablation-acc-byLocalizer-part2}}
\end{figure*}
\begin{table*}[]
\centering
\begin{tabular}{l|c|c|c|c|c|c}
Localizer & \textbf{P1.1} &\textbf{P1.2} & \textbf{P2.1} & \textbf{P2.2} & \textbf{P3.1} & \textbf{P3.2 }\\ \hline
GB & 0.59 & 0.49 & 0.58 & 0.50 & 0.36 & 0.22 \\
(simple) & [0.46, 0.73] & [0.36, 0.61] & [0.46, 0.7] & [0.39, 0.63] & [0.06, 0.64] & [-0.09, 0.55] \\ \hline
MI & 0.19 & -0.11 & 0.16 & -0.08 & 0.08 & 0.08 \\
(simple) & [0.05, 0.32] & [-0.24, 0.02] & [0.04, 0.30] & [-0.20, 0.05] & [-0.22, 0.38] & [-0.24, 0.41] \\ \hline
LB + CI & 0.14 & -0.04 & 0.21 & -0.10 & 0.04 & 0.17 \\ 
(conj.) & [0.01, 0.28] & [-0.17, 0.09] & [0.08, 0.34] & [-0.22, 0.03] & [-0.26, 0.33] & [-0.14, 0.50] \\ \hline
All & 0.22 & -0.05 & 0.29 & 0.05 & 0.10 & 0.19 \\
(simple) & [0.09, 0.36] & [-0.19, 0.08] & [0.17, 0.42] & [-0.07, 0.17] & [-0.19, 0.40]  & [-0.14, 0.50] \\ \hline
CI & 0.08 & -0.12 & 0.16 & -0.13 & 0.03 & 0.13 \\
(conj.) & [-0.05, 0.22] & [-0.24, 0.02] & [0.04, 0.29] & [-0.26, -0.01]  & [-0.27, 0.33] & [-0.19, 0.45] \\ \hline
CI & 0.14 & -0.05 & 0.23 & -0.05 & 0.03 & 0.20 \\
(simple) & [0.01, 0.28] & [-0.19, 0.08] & [0.10, 0.36] & [-0.18, 0.07] & [-0.26, 0.32] & [-0.12, 0.52] \\ \hline
LB & 0.26 & 0.14 & 0.36 & 0.15 & 0.18 & 0.18 \\
(conj.) & [0.13, 0.40] & [0.00, 0.27] & [0.23, 0.48] & [0.02, 0.28] & [-0.12, 0.47] & [-0.14, 0.50] \\ \hline 
LB & 0.34 & 0.27 & 0.39 & 0.22 & 0.20 & 0.19 \\
(simple) & [0.20, 0.47] & [0.14, 0.40] & [0.26, 0.52] & [0.09, 0.34] & [-0.10, 0.50] & [-0.13, 0.51] \\ \hline
\end{tabular}
\caption{Posterior mean estimates and [95\% credible intervals] for assessing the main predictions separately by-localizer. When the credible interval excludes zero, the prediction is borne out. ``conj.'' stands for conjunctive localizers. 
\label{app:tab:byLoc-hypotheses}}
\end{table*}

\subsection{Subnetwork Ablation Results on Additional Datasets}
\label{app:sec:ablations-g-laoding-datasets}
Next to assessing the effect of the subnetwork ablation on ToM and pragmatics performance in comparison to general linguistic ability benchmarks, we evaluated 13 models on additional cognitive skills that may be recruited in ToM and pragmatic tasks, namely on the entity tracking and analogical reasoning benchmarks.

We cast both benchmarks into multiple choice format. 
For the entity tracking benchmark from \citet{kim2023entity}, we use the items from the test dataset from their experiment 1, and convert the masked item format into the following sentence completion prompt:\newline

\noindent \textit{Given the description after ``Description:'', write a true statement about all boxes and their contents according to the description after ``Statement:''. Format the statement as Box 0 contains the A, Box 1 contains the B, Box 2 contains the C, Box 3 contains the D, Box 4 contains the E, Box 5 contains the F, Box 6 contains the G. A, B, C, D, E, F, G are placeholders for the contents of each box.}\newline\newline

\noindent \textit{Description: Box 0 contains \{A\}, Box 1 contains \{B\}, \dots \newline
Statement: Box \{X\} contains }\newline
The performance is assessed by retrieving the conditional probability of each of the $m$ different answer options under the LM, where $m$ is the number of boxes described in the item, one of which described the correct content of the target box $X$.
We match the size of the linguistic benchmarks and sample 670 items from the original dataset. 

For analogical reasoning, the multiple choice formatted verbal analogy and story analogy tasks from \citet{webb2023emergent} were selected.
For verbal analogy, the task is to identify a correct completion which is analogous to an example pair (e.g., given ``clothing : jacket :: bird : '', select among the options pigeon, dog).
LMs are evaluated by scoring each of the possible completions, given the example and first member of the target pair.
For story analogy, the task is to decide, given a story, which of two additional stories is a better analogy to the first story, or if they are equally analogous.
We largely build on the prompts from \citet{webb2023emergent}.

\begin{figure*}
\centering
\includegraphics[width=\textwidth]{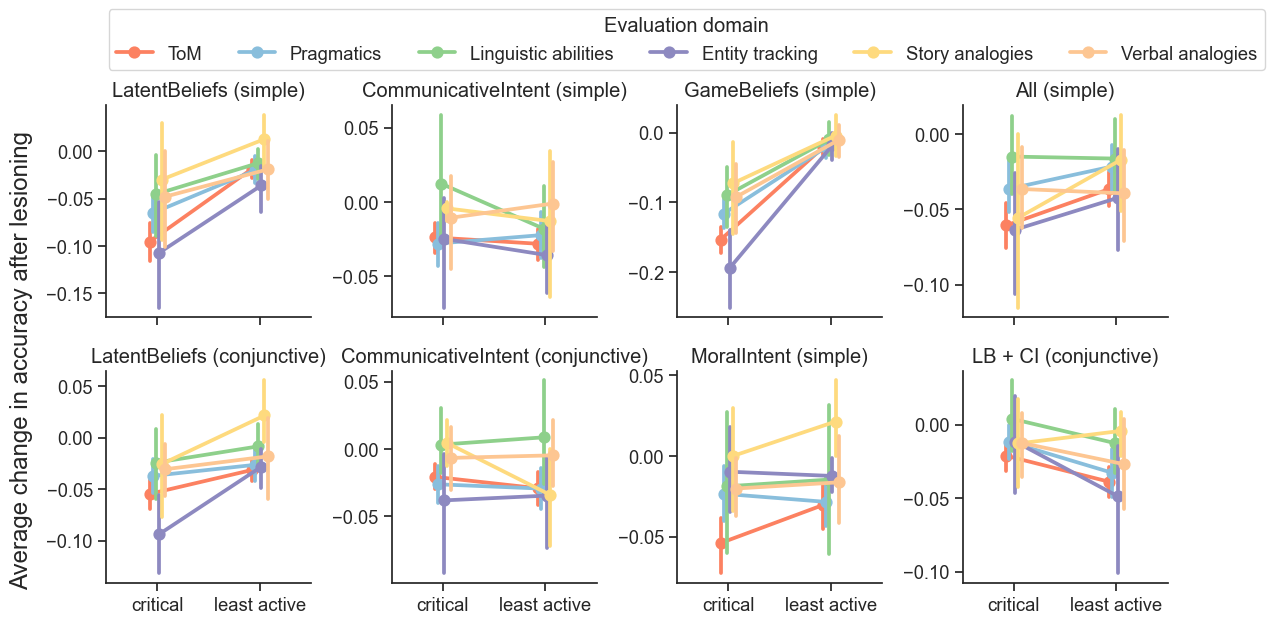}
\caption{Average difference in accuracy under ablations of the localized subnetworks, by localizer, across 13 models. Additionally to the results reported in the main text, change in performance on general reasoning tasks (entity tracking \citealp{kim2023entity}; verbal and story analogies \citealp{webb2023emergent}) is shown.
\label{app:fig:loc-ablation-acc-additional-controls}}
\end{figure*}

\begin{figure*}
\centering
\includegraphics[width=\textwidth]{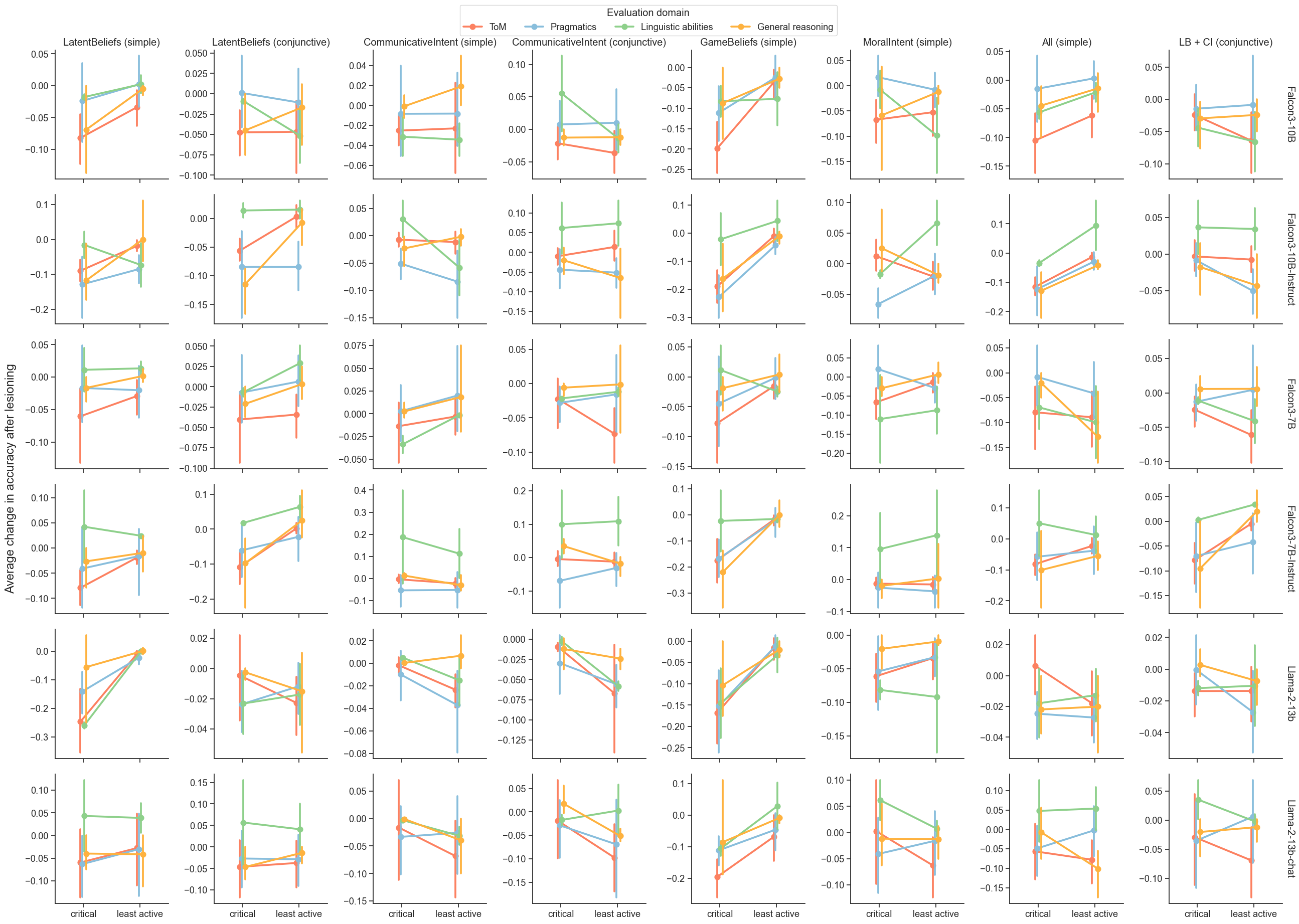}
\caption{Average difference in accuracy under ablations of the localized subnetworks, for each localizer and selected six models. Additionally to the results reported in the main text, average change in performance on control general reasoning tasks is shown.
\label{app:fig:loc-ablation-acc-additional-controls-byModel-part1}}
\end{figure*}

\begin{figure*}
\centering
\includegraphics[width=\textwidth]{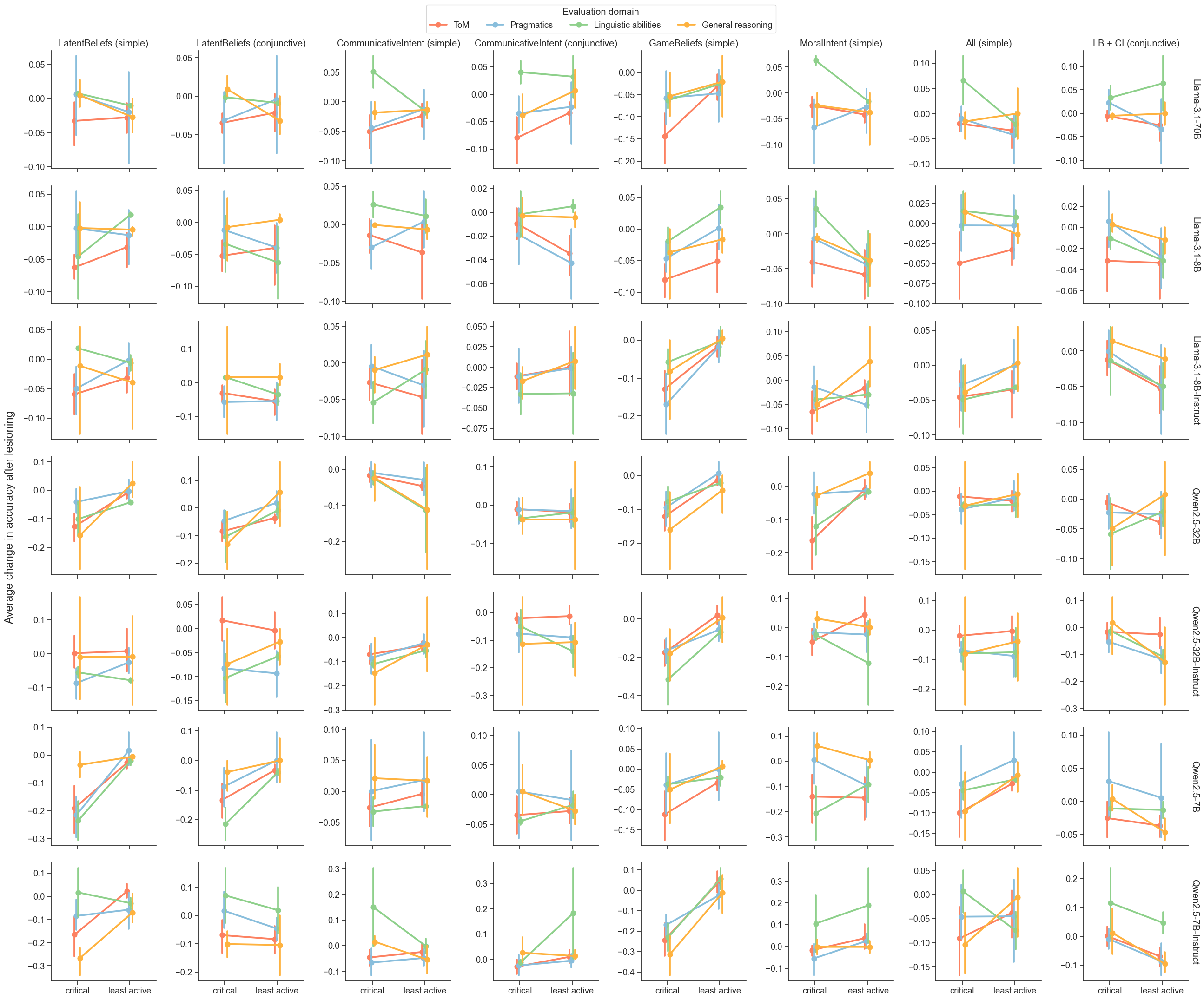}
\caption{Average difference in accuracy under ablations of the localized subnetworks, for each localizer and additional selected seven models. Additionally to the results reported in the main text, average change in performance on control general reasoning tasks is shown.
\label{app:fig:loc-ablation-acc-additional-controls-byModel-part2}}
\end{figure*}

We evaluate the following models on these three datasets: Falcon3-7B, Falcon3-7B-Instruct, Falcon3-10B, Falcon3-10B-Instruct, Llama-3.1-8B, Llama-3.1-8B-Instruct, Llama-2-13b-hf, Llama-2-13b-chat-hf, Llama-3.1-70B, Qwen2.5-7B, Qwen2.5-7B-Instruct, Qwen2.5-32B, Qwen2.5-32B-Instruct.
Each model is evaluated without interventions as well as under ablations of the critical and least active networks localized by each of our localizers. 
Results for each of the localizers, aggregated across models are shown in~Figure~\ref{app:fig:loc-ablation-acc-additional-controls}.
For most localizer suites, including the LatentBeliefs (simple \& conjunctive) and GameBeliefs suites which mainly supported our overall results reported in Section~\ref{sec:ablation-experiments}, the causal effect of the critical subnetwork ablation was largest for the performance on entity tracking.
For analogical reasoning, the effect was smaller (e.g. for GameBeliefs) or qualitatively different (e.g., for CommunicativeIntent conjunctive) than for the critical evaluation domains pragmatics and ToM.
This suggests that ToM and pragmatic capabilities might indeed also rely on shared entity tracking mechanisms, in addition to social reasoning mechanisms, and that these may share representations.
In contrast, analogical reasoning seems to be less likely to be part of ToM and pragmatics tasks.
Results for all 13 models for each localizer are shown in Figures~\ref{app:fig:loc-ablation-acc-additional-controls-byModel-part1}--\ref{app:fig:loc-ablation-acc-additional-controls-byModel-part2} (aggregating the three reasoning datasets into the ``General reasoning'' domain).

\end{document}